\documentclass[letterpaper, 10 pt, journal, final]{IEEEtran}

\usepackage{amsmath} %
\usepackage{amssymb} %
\usepackage{amsfonts}
\usepackage{makecell}
\usepackage{varwidth}

\usepackage[dvipsnames, table]{xcolor}
\usepackage{graphicx}
\usepackage{tabularx}
\usepackage{multirow}
\usepackage{mathtools}
\usepackage{esvect} %
\usepackage{siunitx}
\usepackage{caption}
\usepackage{cite}
\usepackage{flushend} %
\usepackage[pdftex, pdfstartview={FitV}, pdfpagelayout={TwoColumnLeft},bookmarksopen=true,plainpages = false, colorlinks=true, linkcolor=black, citecolor = black, urlcolor = black,filecolor=black , pagebackref=false,hypertexnames=false, plainpages=false, pdfpagelabels ]{hyperref}
\usepackage[T1]{fontenc} %
\usepackage{mathtools, cuted} %

\usepackage{balance} %
\usepackage{booktabs} %
\usepackage{array}
\usepackage[export]{adjustbox} %

\newcolumntype{x}[1]{%
>{\raggedleft\hspace{0pt}}p{#1}}%

\usepackage{algorithm}
\usepackage[noend]{algpseudocode} %
\definecolor{commentclr}{RGB}{34, 139, 34}
\newcommand{\algcomment}[1]{{\color{commentclr}\% #1}}

\usepackage{tikz}

\usetikzlibrary{positioning, fit, shapes.geometric, calc,arrows.meta}

\usepackage{cleveref}
\usepackage{subcaption}
\DeclareCaptionLabelSeparator{periodspace}{.\quad}
\usepackage{amsthm}

\theoremstyle{definition}

\usepackage{letltxmacro}
\LetLtxMacro\orgvdots\vdots
\LetLtxMacro\orgddots\ddots

\makeatletter
\DeclareRobustCommand\vdots{%
	\mathpalette\@vdots{}%
}
\newcommand*{\@vdots}[2]{%
	\sbox0{$#1\cdotp\cdotp\cdotp\m@th$}%
	\sbox2{$#1.\m@th$}%
	\vbox{%
		\dimen@=\wd0 %
		\advance\dimen@ -3\ht2 %
		\kern.5\dimen@
		\dimen@=\wd2 %
		\advance\dimen@ -\ht2 %
		\dimen2=\wd0 %
		\advance\dimen2 -\dimen@
		\vbox to \dimen2{%
			\offinterlineskip
			\copy2 \vfill\copy2 \vfill\copy2 %
		}%
	}%
}
\DeclareRobustCommand\ddots{%
	\mathinner{%
		\mathpalette\@ddots{}%
		\mkern\thinmuskip
	}%
}
\newcommand*{\@ddots}[2]{%
	\sbox0{$#1\cdotp\cdotp\cdotp\m@th$}%
	\sbox2{$#1.\m@th$}%
	\vbox{%
		\dimen@=\wd0 %
		\advance\dimen@ -3\ht2 %
		\kern.5\dimen@
		\dimen@=\wd2 %
		\advance\dimen@ -\ht2 %
		\dimen2=\wd0 %
		\advance\dimen2 -\dimen@
		\vbox to \dimen2{%
			\offinterlineskip
			\hbox{$#1\mathpunct{.}\m@th$}%
			\vfill
			\hbox{$#1\mathpunct{\kern\wd2}\mathpunct{.}\m@th$}%
			\vfill
			\hbox{$#1\mathpunct{\kern\wd2}\mathpunct{\kern\wd2}\mathpunct{.}\m@th$}%
		}%
	}%
}
\makeatother

\let\oldnl\nl%
\newcommand{\nonl}{\renewcommand{\nl}{\let\nl\oldnl}}%
\makeatother

\renewcommand{\b}[1]{{\color{blue}{#1}}}

\graphicspath{{figures/}}

\crefname{figure}{Fig.}{Figs.}

\algblockdefx{RetStruct}{EndRetStruct}[1]{\textbf{return} \textbf{struct} \texttt{#1}}{\textbf{end struct}}

\algblockdefx{Struct}{EndStruct}[1]{\textbf{struct} \texttt{#1}}{\textbf{end struct}}

\definecolor{ResultRed}{HTML}{F4CCCC}
\definecolor{ResultYellow}{HTML}{FFF2CC}
\definecolor{ResultGreen}{HTML}{D9EAD3}

\begin{document}

\title{\textsc{PARTE}: Plane-Assisted Robust Transformation Estimation for Point Cloud Registration} 

\author{%
Abolfazl Babanazari$^{1}$, Carson Cramer$^{1}$, Tyler Summers$^{2}$, Carlos Nieto$^{3}$, Kaveh Fathian$^{1}$%
 \thanks{$^{1}$Department of Computer Science, Colorado School of Mines. Email: abolfazl\_babanazari@mines.edu, fathian@ariarobotics.com}
 \thanks{$^{2}$ Department of Mechanical Engineering, University of Texas at Dallas. Email: tsummers@utdallas.edu}
 \thanks{$^{3}$U.S. Army Combat Capabilities Development Command, Army Research Laboratory {Email: carlos.p.nieto2.civ@army.mil}}
}%

\maketitle

\begin{abstract}

Global point-cloud registration remains challenging when limited overlap, repetitive geometry, and sensor noise produce correspondence sets dominated by outliers.
Planar regions are particularly difficult for conventional point descriptors and are therefore often suppressed or discarded before matching.
We present PARTE (Plane-Assisted Robust Transformation Estimation), a global registration method that instead treats planar structure as complementary registration evidence.
PARTE extracts planar patches and represents them using our novel Plane Context Histogram (PCH), a descriptor that encodes the geometry surrounding each patch, while a two-level matching procedure identifies reliable plane correspondences.
Candidate point and plane correspondences are combined in a confidence-weighted compatibility graph for joint outlier rejection, followed by rigid transformation estimation.
When no usable plane correspondences are available, PARTE naturally reduces to point-only registration.
We evaluate PARTE on $8,097$ registration pairs across six indoor and outdoor benchmarks spanning dense RGB-D and sparse LiDAR measurements.
Evaluations show PARTE achieves the highest overall success rate against $13$ standard and state-of-the-art methods while maintaining low runtime.
An open-source C++ implementation with Python bindings is provided at \href{https://ariarobotics.github.io/parte/}{\b{https://ariarobotics.github.io/parte/}}.
\end{abstract}

\section{Introduction}\label{sec:intro}
{

\begin{figure}[!th]
\centering
\includegraphics[width=\columnwidth]{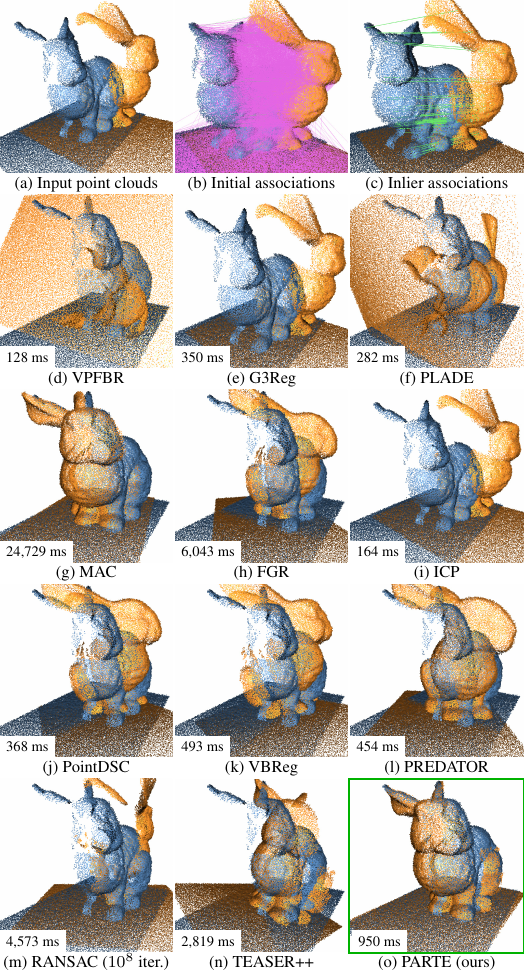}
\caption{ 
Registration of noisy Stanford bunny point clouds augmented with ground-plane points using several state-of-the-art algorithms.
(a) Input point clouds colored blue and orange, with Gaussian noise of $\mathcal{N}(0, 10^{-3})$ added to each point coordinate.
(b) About $17{,}000$ initial associations obtained by nearest-neighbor matching of FPFH point descriptors.
(c) Ground-truth inlier associations, which constitute less than $0.2\%$ of the initial associations.
(d--n) Registered point clouds produced by various classical and state-of-the-art algorithms.
(o) Registered point clouds produced by PARTE.}
\label{fig:intro-fig}

\vspace{-1em}
\end{figure}
}

A critical bottleneck in developing advanced autonomous robots is the ability to create an accurate 3D map of their environment in the presence of sensor noise, dynamic objects, and limited computational resources. 
Robotic mapping systems typically represent each sensor observation as a point cloud, a set of 3D measurements that captures the geometry of the surrounding environment.
As the robot captures successive point clouds while traveling through its environment, the core challenge is determining how the robot moved between scans in order to stitch them into a coherent map.
This alignment process is known as \textit{point cloud registration} and underpins many important applications, including SLAM, object recognition, augmented reality, and medical imaging.

Achieving fast, accurate, and robust point cloud registration remains challenging. 
As illustrated in Fig.~\ref{fig:intro-fig}, achieving both high registration accuracy and low runtime remains difficult for existing methods.
Although globally optimal registration algorithms exist~\cite{GO-ICP}, their exponential runtime scaling limits their use in real-time applications.
Practical registration methods therefore generally follow one of two strategies: local optimization or feature-based global registration.
Local methods iteratively minimize an alignment error between two point clouds. They are fast and accurate when provided with a good initial estimate, but may converge to an incorrect local minimum when the initial misalignment is large.
Feature-based global methods instead search for transformation-invariant features in the two point clouds, establish tentative correspondences, and estimate the transformation without requiring an initial alignment.
This makes them more versatile and particularly useful for tasks such as loop closure, re-localization, and initialization-free scan matching.
However, their reliability is often limited by incorrect correspondences, commonly referred to as outliers.
These outliers arise from multiple sources, including dynamic objects, sensor noise, limited overlap, and feature-poor regions where distinctive geometric patterns are scarce.
Planar regions are particularly challenging for point-based matching because
points sampled from the same or similar planar surfaces often have similar local descriptors, making their correspondences highly ambiguous.
Since planar surfaces are common in robotic environments, these ambiguities can substantially increase the outlier ratio and reduce registration reliability.

Existing global registration methods often treat planar structure as a source of ambiguous correspondences.
Some remove specific planes, most commonly the ground plane, while others discard low-curvature planar points before feature matching.
Although these strategies can reduce ambiguous point correspondences, they also discard large portions of the scene that provide useful structural cues.
Moreover, removing planar points entirely can isolate nearby edge and corner points, leaving too few neighbors for reliable descriptor computation.
The key observation behind PARTE is that, although individual points on a plane may be difficult to distinguish, the planar patch itself can provide a distinctive geometric feature and impose strong constraints on the relative pose.

Based on this observation, we formulate global registration over a mixed set of point and planar-patch correspondences.
PARTE first uses a greedy region-growing algorithm to extract planar patches, while the remaining points are separated for point matching.
For plane matching, we introduce a novel descriptor, named the Plane Context Histogram (PCH), alongside a two-level matching procedure for establishing reliable plane correspondences.
During point-feature computation, planar points are retained as neighbors but excluded as standalone features.
The resulting point and plane correspondences are jointly processed using a graph-theoretic outlier-rejection framework extended to support plane correspondences.
To reduce sensitivity to manually tuned thresholds, PARTE uses geometry-dependent consistency tolerances that account for uncertainty in the matched planes.
Plane correspondences are further weighted by their descriptor confidence, resulting in a weighted maximum-clique formulation.
The remaining point and plane correspondences are then jointly optimized to recover the rigid transformation.

We evaluate PARTE on $8,097$ registration pairs across six benchmarks spanning dense RGB-D and sparse LiDAR measurements, indoor and outdoor environments, and point clouds containing between $10^4$ and $10^6$ points.
Across the four principal robotics benchmarks---3DMatch, 3DLoMatch, KITTI-10m, and KITTI-LC---PARTE achieves the highest mean success rate among the evaluated methods while maintaining low end-to-end runtime.
PARTE also achieves the highest overall success rate on both additional benchmarks: ETH and the partial-to-full RESSO benchmark.

In summary, the main contributions of this work are
\begin{itemize}
\item We present PARTE, a plane-assisted global registration pipeline that jointly uses point and planar-patch correspondences in graph-based outlier rejection and rigid-transformation estimation.

\item We introduce the Plane Context Histogram, a descriptor that represents a planar patch using its surrounding geometry, outperforming $14$ existing shape descriptor and plane matching algorithms.

\item We formulate joint point--plane outlier rejection as a confidence-weighted maximum-clique problem, using geometry-dependent compatibility constraints that account for uncertainty in the extracted planes.

\item We evaluate PARTE on $8,097$ registration pairs across six indoor and outdoor benchmarks. PARTE achieves the highest mean success rate across the four principal benchmarks and the highest overall success rate on both ETH and partial-to-full RESSO.
\end{itemize}

\section{Related Work}\label{sec:related}
\textbf{Feature-based global registration.}
Most correspondence-based global registration methods represent a point cloud using a subset of distinctive point features.
These features may be selected using hand-crafted keypoint detectors~\cite{sipiran2011harris,zhong2009intrinsic,steder2010narf} or learned directly from data~\cite{li2019usip,bai2020d3feat}.
Local descriptors are then computed around the selected points and matched between scans.
Classical descriptors include FPFH~\cite{rusu2009fast}, SHOT~\cite{tombari2010shot}, Spin Images~\cite{johnson1997spin}, and 3D Shape Context~\cite{kortgen20033d}, while more recent methods use learned features and matching networks~\cite{zeng20173dmatch,choy2019fully,bai2020d3feat,huang2021predator,ao2021spinnet,qin2023geotransformer}.
Because repetitive or weakly structured regions can produce ambiguous point features, some methods additionally suppress such regions before matching.
For example, QUATRO++ removes the ground plane before registration~\cite{lim2024quatropp}, while KISS-Matcher suppresses low-variation geometric regions that tend to produce ambiguous features~\cite{lim2024kiss}.

\textbf{Structure-aware registration.}
An alternative is to explicitly extract larger geometric or semantic structures and use them as registration primitives.
The use of geometric primitives for 3D alignment dates back to early work on representing scenes using lines, planes, and other geometric entities~\cite{faugeras19833,faugeras1986representation}.
Later methods developed robust plane and line extraction for LiDAR data~\cite{brenner2007automatic,von2008line,tao2020fast} and incorporated planar constraints directly into registration.
Point-to-plane ICP~\cite{chen1992object}, GP-ICP~\cite{kim2019gp}, and NAP-ICP~\cite{favre2021plane}, for example, use planar structure within local ICP-style alignment.
More recent global methods employ richer geometric primitives: G3Reg~\cite{qiao2024g3reg} uses geometric structures, and  QuadricsReg~\cite{wu2024quadricsreg} represents scene elements using quadric primitives and optionally combines their semantic labels with intrinsic geometric attributes.
Semantic registration methods similarly replace or augment individual point matches with higher-level scene information.
Examples include SE-NDT~\cite{zaganidis2017semantic}, SegICP~\cite{wong2017segicp}, SegReg~\cite{mei2022partial}, SemReg~\cite{fung2024semreg}, and SarNet~\cite{qin2024sarnet}.
Semantic approaches can provide strong priors in repetitive environments, but typically require semantic prediction or learned representations in addition to the geometric registration pipeline.

\textbf{Geometric descriptors and plane matching.}
Planes have been widely used as geometric features in registration and SLAM.
In local registration, plane correspondences can often be established using a prior pose estimate or the small motion between consecutive scans, and simple plane properties such as orientation, position, extent, or overlap are often sufficient for association.
This allows methods such as Plane-ICP, GP-ICP~\cite{kim2019gp}, and NAP-ICP~\cite{favre2021plane}, as well as plane-based SLAM systems such as Linear RGB-D SLAM~\cite{kim2018linear}, PlanarSLAM~\cite{li2021rgbd}, and VIP-SLAM~\cite{chen2022vipslam}, to use planar features as additional constraints during pose estimation.

For global plane matching, simple geometric properties such as plane orientation, centroid, size, or boundary shape can be used to identify candidate correspondences, but these properties are often ambiguous when several similar planes are present in the same scene.
Several methods therefore incorporate richer information.
HIA-TCD~\cite{wei2022plane} describes planar patches using their contour geometry, while PLADE~\cite{chen2019plade} uses relationships between planes and lines to establish structure-level correspondences.
Plane Pair Matching~\cite{kaiser2020planepair} similarly exploits relationships between pairs of planes rather than describing each patch independently.
Other approaches learn the representation directly from data; PlaneMatch~\cite{shi2018planematch}, for example, learns an RGB-D descriptor for matching coplanar patches across different views.

Generic 3D shape descriptors provide another way to represent planar patches.
Descriptors such as ESF~\cite{wohlkinger2011esf}, Spin Images~\cite{johnson1997spin}, 3D Shape Context~\cite{kortgen20033d}, GASD~\cite{lima2016gasd}, and the handcrafted features used by SegMatch~\cite{dube2017segmatch} encode the shape of an object or segment through distributions of point positions, surface orientations, or simple geometric statistics.
Although these descriptors can be applied to planar patches, they were developed primarily for object, surface, or segment recognition rather than for distinguishing similar planes across different scenes.
In contrast, PCH describes a planar patch using the geometry surrounding the plane, providing additional context when the shape of the patch itself is ambiguous.

\textbf{Robust correspondence filtering.}
Feature matching often produces a large number of incorrect correspondences, making robust outlier rejection an important stage of registration.
RANSAC~\cite{fischler1981random} and its many variants~\cite{chum2005matching,chum2003locally,martinez2022ransac,
hu2022daniel,sun2021ransic,shi2024ransac} address this problem by repeatedly sampling small correspondence subsets and selecting transformations supported by a large set.
Other approaches estimate the transformation using robust optimization~\cite{zhou2016fast,chebrolu2021adaptive,yang2020graduated,chen2024fracgm,yang2020teaser}.
Learning-based methods instead learn to identify or reweight reliable correspondences, including 3DRegNet~\cite{pais20203dregnet}, PointDSC~\cite{bai2021pointdsc}, VBReg~\cite{jiang2023robust}, and BUFFER~\cite{ao2023buffer}.

Graph-based methods have recently become particularly attractive for registration with very high outlier ratios.
They represent candidate correspondences as graph nodes and connect geometrically consistent pairs, allowing mutually compatible correspondences to be identified without first estimating the transformation.
Methods such as ROBIN~\cite{shi2021robin}, TEASER++~\cite{yang2020teaser}, CLIPPER~\cite{lusk2021clipper}, SC$^2$-PCR~\cite{chen2022sc2}, and CLIPPER+~\cite{fathian2024clipper+} build on this idea using compatibility graphs and clique-based formulations.
A major advantage of this family, which PARTE also exploits, is its robustness to extreme outlier ratios, in some cases as high as $99\%$.

PARTE builds on graph-based global registration by introducing plane correspondences as complementary evidence to conventional point matches. 
PARTE's plane matching also targets planar regions in which local point descriptors are ambiguous by representing the geometry surrounding each planar patch.
These correspondences are integrated with point matches throughout compatibility testing, confidence-weighted inlier selection, and transformation estimation.

\begin{figure*}[t]
    \centering
    \includegraphics[width=\textwidth]{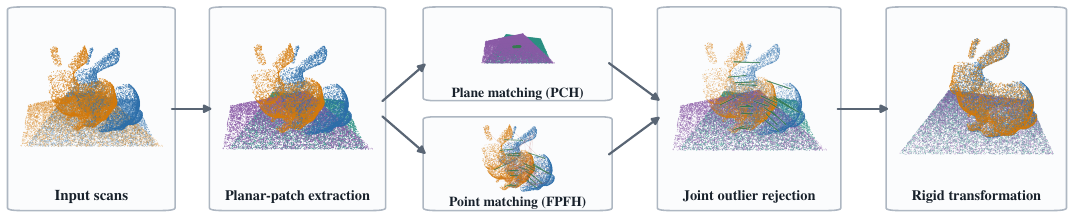}
    \caption{Overview of the PARTE registration pipeline. Planar patches are extracted from the input scans and matched by PCH, while the remaining points are matched by FPFH. Candidate plane and point correspondences are combined for joint outlier rejection and rigid-transformation estimation.}
    \label{fig:parte-pipeline-preview}
\end{figure*}

\section{Notation}
The main notation used throughout the paper is summarized in Table~\ref{tab:notation}.
For a finite set $\mathcal X$, $|\mathcal X|$ denotes its cardinality.
For a vector $x\in\mathbb R^n$, we use $\|x\|_1=\sum_i |x_i|$ and $\|x\|_2=(\sum_i x_i^2)^{1/2}$ to denote the $\ell_1$ and $\ell_2$ norms, respectively.
For nonnegative vectors $h,g\in\mathbb R^n$, we define their $\chi^2$-distance as
\begin{equation}
\chi^2(h,g)=\sum_{i:h_i+g_i>0}\frac{(h_i-g_i)^2}{h_i+g_i}.
\end{equation}

\begin{table}[t]
\centering
\caption{Notation used throughout the paper.}
\label{tab:notation}
\small
\begin{tabularx}{\columnwidth}{@{}lX@{}}
\toprule
Symbol & Description \\
\midrule
\multicolumn{2}{@{}l}{\textbf{General}} \\[1mm]
$v\in\mathbb R_{>0}$ & Voxel-downsampling size. \\
$R\in\mathrm{SO}(3)$ & Rotation between the two point clouds. \\
$t\in\mathbb R^3$ & Translation between the two point clouds. \\
$\mathbb S^2$ & Unit sphere in $\mathbb R^3$. \\
$\mathrm{SO}(3)$ & Group of 3D rotation matrices. \\

\addlinespace
\multicolumn{2}{@{}l}{\textbf{Point clouds}} \\[1mm]
$P,Q\subset\mathbb R^3$ & Source and target point clouds. \\
$p_i\in P,\;q_i\in Q$ & Source and target points. \\
$\nu_i\in\mathbb S^2$ & Estimated unit normal of point $p_i$. \\

\addlinespace
\multicolumn{2}{@{}l}{\textbf{Plane segmentation}} \\[1mm]
$\tau_d$ & Plane-thickness threshold. \\
$\tau_\theta$ & Plane-normal dispersion threshold. \\

\multicolumn{2}{@{}l}{\textbf{Planar patches}} \\[1mm]
$\Pi=\{\pi_j\}$ & Set of extracted planar patches. \\
$\mathcal I_j$ & Indices of points belonging to patch $\pi_j$. \\
$\mu_j\in\mathbb R^3$ & Centroid of patch $\pi_j$. \\
$u_j\in\mathbb S^2$ & Estimated unit normal of patch $\pi_j$. \\
$d_j\in\mathbb R$ & Plane offset, $d_j=u_j^\top\mu_j$. \\

\addlinespace
\multicolumn{2}{@{}l}{\textbf{Plane descriptors}} \\[1mm]
$r_{\mathrm{PCH}}\in\mathbb R_{>0}$ & PCH signed-distance half-range. \\

\addlinespace
\multicolumn{2}{@{}l}{\textbf{Correspondences}} \\[1mm]
$\mathcal C_{\mathrm{pt}}=\{(p_i,q_i)\}$ & Candidate point correspondences. \\
$\mathcal C_{\mathrm{pl}}=\{(\pi_j^P,\pi_j^Q)\}$ & Candidate plane correspondences. \\
$\widehat{\mathcal C}_{\mathrm{pt}}$ & Inlier point correspondences. \\
$\widehat{\mathcal C}_{\mathrm{pl}}$ & Inlier plane correspondences. \\
$w_j\in[0,1]$ & Confidence of plane correspondence $j$. \\

\addlinespace
\multicolumn{2}{@{}l}{\textbf{Outlier rejection}} \\[1mm]
$\epsilon_p$ & Base spatial consistency tolerance. \\
$\epsilon_\theta$ & Base angular consistency tolerance. \\
\bottomrule
\end{tabularx}
\end{table}

Superscripts $P$ and $Q$ identify quantities associated with the source and target point clouds, respectively. For example, $\pi_j^P$ and $\pi_j^Q$ denote a pair of matched planar patches from the two clouds.
For notational simplicity, matched features are re-indexed by correspondence. Therefore, $(p_i,q_i)$ denotes the $i$th point correspondence, while
$(\pi_j^P,\pi_j^Q)$ denotes the $j$th plane correspondence.
For a symmetric matrix $A\in\mathbb{R}^{3\times3}$, we denote its eigenvalues and corresponding unit eigenvectors by $\lambda_i(A)$ and $v_i(A)$, ordered as
\begin{equation}
\lambda_1(A)\geq\lambda_2(A)\geq\lambda_3(A),
\qquad
Av_i(A)=\lambda_i(A)v_i(A).
\end{equation}

\section{Proposed Approach}\label{sec:approach}
PARTE follows the standard feature-based registration stages of feature extraction, descriptor matching, outlier rejection, and transformation estimation, as illustrated in Fig.~\ref{fig:parte-pipeline-preview}.
Its distinction lies in how planar information is represented and integrated throughout these stages.
PARTE describes each extracted planar patch through the surrounding scene geometry using PCH, while retaining conventional FPFH point features for non-planar regions.
The resulting point and plane correspondences are combined in a single compatibility graph using mixed point--point, plane--plane, and plane--point consistency tests.
Descriptor confidence controls the influence of each plane correspondence during maximum-clique selection, and the retained mixed correspondences jointly constrain the final rigid transformation.
When no suitable planar patches are extracted, the same formulation reduces to point-only registration.

\subsection{Preprocessing and Feature Extraction}
\label{sec:preprocessing}
Preprocessing starts by downsampling each point cloud to reduce data size and the computational cost of registration.
Surface normals are then estimated for the downsampled points.
The downsampled points are segmented into planar patches, while the remaining points are retained as point features.
Together, the planar patches and point features are used for joint plane--point registration in the subsequent stages.

\textbf{Downsampling.}
We use voxel-based downsampling with voxel size $v$.
Three-dimensional space is divided into cubic voxels of edge length $v$, and the points within each occupied voxel are replaced by their centroid.

\textbf{Normal estimation.}
For each downsampled point $p_i$, we estimate a unit surface normal $\nu_i$
from its $k$-nearest-neighbor neighborhood using local PCA.
The normal is given by the eigenvector corresponding to the smallest eigenvalue of the neighborhood covariance matrix.
We consistently orient the estimated normals toward the sensor viewpoint.

\begin{algorithm}[th]
\caption{Planar Patch Extraction}
\label{alg:planar-patch-extraction}
\small
\begin{algorithmic}[1]
\State \textbf{Input} Point cloud $P=\{p_i\}_{i=1}^{N}$ and point normals $\{\nu_i\}_{i=1}^{N}$.
\State \textbf{Parameters} Thickness threshold $\tau_d$, normal-dispersion threshold $\tau_\theta$, and minimum patch size $\mathrm{min\_pts}$.
\State \textbf{Output} $\Pi = \{ \pi_j \}$ \hfill \algcomment{Set of planar patches}

\Statex
\State \algcomment{Planar seed initialization}
\ForAll{$p_i\in P$}
    \State $\bar C_i,\bar Q_i\gets$ neighborhood statistics of $p_i$ from ~\eqref{eq:patch-point-statistics} and~\eqref{eq:patch-normal-moment}
    \If{$\lambda_3(\bar C_i)\leq\tau^2_d$ \textbf{ and } $1-\lambda_1(\bar Q_i)\leq\tau^2_\theta$}
        \State $\pi_i\gets$ a patch initialized from $(p_i,\nu_i)$
        \State $\Pi\gets\Pi\cup\{\pi_i\}$
    \EndIf
\EndFor
\Statex
\State \algcomment{Planar patch merging}
\Repeat
    \ForAll{neighboring pairs $(\pi_a,\pi_b)\in\Pi$}
        \State $\mathcal I_{c},\mu_{c},C_{c},Q_{c}
        \gets$ merged patch statistics
        \State $u_{c}\gets$ solution of ~\eqref{eq:patch-normal}
        \If{$u_{c}^{\top}C_{c}u_{c}\leq\tau^2_d$
            \textbf{ and }
            $1-u_{c}^{\top}Q_{c}u_{c}\leq\tau^2_\theta$}
            \State $\pi_{c}\gets(\mathcal I_{c},\mu_{c},C_{c},Q_{c},u_{c})$
            \State $\Pi\gets
            \left(\Pi\setminus\{\pi_a,\pi_b\}\right)\cup\{\pi_{c}\}$
        \EndIf
    \EndFor
\Until{\textit{no patches are merged}}

\Statex
\State \algcomment{Remove patches with insufficient support}
\State $\Pi\gets$ remove $\pi_j$ with
$\lvert\mathcal I_j\rvert<\mathrm{min\_pts}$
    
\State \Return $\Pi$
\end{algorithmic}
\end{algorithm}

\textbf{Planar-patch segmentation.}
We segment the input point cloud into planar regions using a region-growing strategy.
The algorithm assumes that the input point cloud ${P=\{p_1,\ldots,p_N\}}$ has an estimated normal $\nu_i\in\mathbb{S}^2$ associated with each point $p_i$.
At each stage, the algorithm maintains a set of planar patches $\Pi=\{\pi_j\}$, where each patch $\pi_j$ has a support set $\mathcal{I}_j\subseteq\{1,\ldots,N\}$.
As shown in Lines~4--9 of Algorithm~\ref{alg:planar-patch-extraction}, the algorithm initializes a set of planar patches, each containing one point whose local neighborhood is planar.
It then repeatedly merges neighboring patches whenever their union remains planar (Lines~10--18).
The process terminates when no further valid merge exists, after which patches with insufficient support are discarded.

To determine whether a patch $\pi_j$ remains planar, let $\mu_j$ and $u_j\in\mathbb{S}^2$ denote the centroid and estimated unit normal of its support points $\{p_i : i\in\mathcal I_j\}$.
Given user-selected thresholds $\tau_d$ and $\tau_\theta$, we require every patch $\pi_j\in\Pi$ to satisfy
\begin{align}
    \frac{1}{|\mathcal{I}_j|} \sum_{i\in\mathcal{I}_j} \left(u_j^\top(p_i-\mu_j)\right)^2 &\leq \tau^2_d, \\
    \frac{1}{|\mathcal{I}_j|} \sum_{i\in\mathcal{I}_j} \left[1-(u_j^\top\nu_i)^2\right] &\leq \tau^2_\theta.
\end{align}
The first quantity is the mean squared distance of the support points to the estimated plane.
The second measures the average disagreement between the patch normal and the normals of its support points.

Directly evaluating these conditions would require iterating over all support points after each candidate merge.
Instead, for each patch $\pi_j$, we maintain the centroid and covariance of its support points,
\begin{equation}
\label{eq:patch-point-statistics}
\mu_j = \frac{1}{|\mathcal I_j|} \sum_{i\in\mathcal I_j} p_i,
\qquad
C_j = \frac{1}{|\mathcal I_j|} \sum_{i\in\mathcal I_j}(p_i-\mu_j)(p_i-\mu_j)^\top .
\end{equation}
We also maintain the second-moment matrix of the corresponding point normals,
\begin{equation}
\label{eq:patch-normal-moment}
Q_j = \frac{1}{|\mathcal I_j|} \sum_{i\in\mathcal I_j} \nu_i\nu_i^\top .
\end{equation}
Using ~\eqref{eq:patch-point-statistics} and \eqref{eq:patch-normal-moment}, the planarity conditions reduce to
\begin{equation}
\label{eq:patch-validity}
u_j^\top C_j u_j \leq \tau^2_d, \qquad 1-u_j^\top Q_j u_j \leq \tau^2_\theta .
\end{equation}

\begin{figure}[!t]
  \centering

  \begin{subfigure}[b]{0.48\columnwidth}
    \centering
    \includegraphics[width=\linewidth]{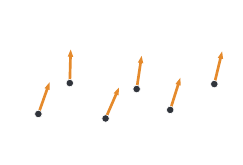}
    \caption{Input points and normals}
    \label{fig:normal-selection-compact-input}
  \end{subfigure}
  \hfill
  \begin{subfigure}[b]{0.48\columnwidth}
    \centering
    \includegraphics[width=\linewidth]{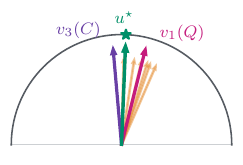}
    \caption{Point and estimated patch normals}
    \label{fig:normal-selection-compact-directions}
  \end{subfigure}
  \begin{subfigure}[b]{0.48\columnwidth}
    \centering
    \includegraphics[width=\linewidth]{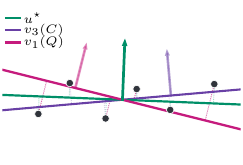}
    \caption{Point residuals of the estimated planes}
    \label{fig:normal-selection-compact-distances}
  \end{subfigure}
  \hfill
  \begin{subfigure}[b]{0.48\columnwidth}
    \centering
    \includegraphics[width=\linewidth]{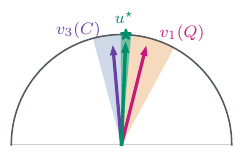}
    \caption{Feasibility intervals}
    \label{fig:normal-selection-compact-feasible}
  \end{subfigure}

    \caption{
        Illustration of normal estimation.
        (a) Six noisy points and their normals.
        (b) The point-based estimate $v_3(C_j)$ can be inconsistent with the point normals.
        (c) Conversely, estimating the patch normal from the individual point normals using $v_1(Q_j)$ can produce large point-to-plane residuals.
        (d) The selected direction $u^\star$ satisfies both planarity conditions.
    }
  \label{fig:normal-selection-compact}
  \label{fig:normal-selection}
\end{figure}
The choice of patch normal $u_j$ affects both planarity conditions.
It can be estimated either from the covariance of the support points, using $v_3(C_j)$ to minimize the patch thickness, or from their point normals, using $v_1(Q_j)$ to maximize normal agreement. 
However, neither estimate alone guarantees that both planarity conditions are satisfied.
Fig.~\ref{fig:normal-selection-compact} illustrates this through an extreme two-dimensional example, where the covariance-based estimate fits the points well but is not consistent with the point normals, while the normal-based estimate agrees with the point normals but is not consistent with the points.
We therefore select, among all unit normals that are consistent with the points, the direction that best agrees with the point normals
\begin{equation}
\label{eq:patch-normal}
u_j =
\underset{\|u\|_2=1}{\arg\min}
\left(1-u^\top Q_j u\right)
\quad
\text{s.t.} 
\quad 
u^\top C_j u \leq \tau^2_d .
\end{equation}
The resulting three-dimensional problem is solved efficiently through its one-dimensional KKT dual.

Algorithm~\ref{alg:planar-patch-extraction} applies these planarity tests during both initialization and region growing.
During initialization, each point whose $k$-nearest-neighbor neighborhood satisfies \eqref{eq:patch-validity} initializes a singleton patch. All other points are excluded from the region-growing process.
During region growing, neighboring patches $\pi_a$ and $\pi_b$ form a candidate patch $\pi_{c}$ with support $\mathcal I_{c}=\mathcal I_a\cup\mathcal I_b$.
Its statistics are computed from those of the two patches, and its normal is estimated using \eqref{eq:patch-normal}.
The merge is finalized only if $\pi_{c}$ satisfies \eqref{eq:patch-validity}.
The process continues until no valid merge remains, after which patches with fewer than $\mathrm{min\_pts}$ points are discarded.

\subsection{Descriptor generation and matching}
In this stage, we compute descriptors for the non-planar point features and extracted planar patches, and independently establish point and plane correspondences between the source and target point clouds.

\begin{figure}[t]
\centering
  \begin{subfigure}[b]{0.55\columnwidth}
    \centering
    \includegraphics[width=\linewidth]{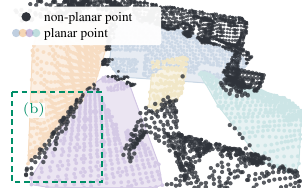}
    \caption{FPFH descriptor centers}
    \label{fig:fpfh-centers}
  \end{subfigure}
  \hfill
  \begin{subfigure}[b]{0.43\columnwidth}
    \centering
    \includegraphics[width=\linewidth]{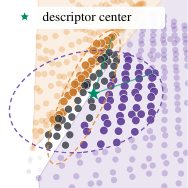}
    \caption{Neighborhood support}
    \label{fig:fpfh-neighborhood}
  \end{subfigure}
    \caption{
      FPFH descriptor computation after plane extraction.
      (a) Descriptors are calculated only at non-planar points.
      (b) Planar points are excluded as descriptor centers but kept in the
      neighborhood used for descriptor computation.
    }
  \label{fig:fpfh-support}
\end{figure}

\textbf{Point descriptor and matching.}
We use FPFH~\cite{rusu2009fast} to compute a $33$-dimensional descriptor for each non-planar point.
FPFH captures the local geometric structure around a point using the relative positions and normals of its neighbors.
As illustrated in Fig.~\ref{fig:fpfh-support}, planar points are excluded as descriptor centers to avoid ambiguous matches, but remain in the neighborhoods of nearby non-planar points during descriptor computation.
We match points whose descriptors are mutual nearest neighbors under the $\ell_2$ distance, producing the candidate point-correspondence set $\mathcal C_{\mathrm{pt}}=\{(p_i,q_i)\}_{i=1}^{N_{\mathrm{pt}}}$.
Although FPFH often produces a high fraction of incorrect correspondences, PARTE's graph-based outlier-rejection stage is designed to tolerate such cases.
This allows the pipeline to use an efficient CPU-based descriptor without relying on more computationally expensive learned alternatives.
The point-descriptor stage is modular and can be replaced by other local descriptors without modifying the remaining pipeline.

\begin{figure}[!t]
  \centering
  \begin{subfigure}[b]{0.48\columnwidth}
    \centering
    \includegraphics{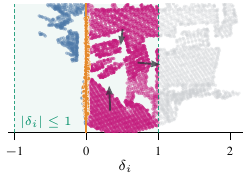}
    \caption{Descriptor support}
    \label{fig:pch-construction-neighborhood}
  \end{subfigure}
  \hfill
  \begin{subfigure}[b]{0.48\columnwidth}
    \centering
    \includegraphics{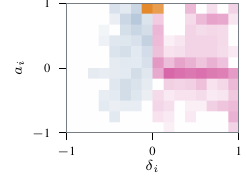}
    \caption{Plane Context Histogram}
    \label{fig:pch-construction-joint}
  \end{subfigure}
  \begin{subfigure}[b]{0.48\columnwidth}
    \centering
    \includegraphics{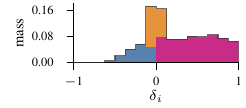}
    \caption{}
    \label{fig:pch-construction-distance}
  \end{subfigure}
  \hfill
  \begin{subfigure}[b]{0.48\columnwidth}
    \centering
    \includegraphics{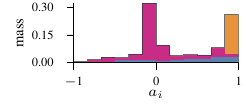}
    \caption{}
    \label{fig:pch-construction-agreement}
  \end{subfigure}
  \caption{
    Construction of the Plane Context Histogram.
    (a) Points within $|\delta_i|\leq r_{\mathrm{PCH}}$ of the reference plane are marked. Orange denotes points of the patch, while blue and magenta denote nearby context points on opposite sides of the patch.
    (b) Each point contributes a signed-distance and normal-agreement pair $(\delta_i,a_i)$, which is accumulated into the normalized two-dimensional PCH.
    (c,d) Signed-distance and normal-agreement distributions.
  }
  \label{fig:pch-construction}
\end{figure}

\textbf{Plane descriptor and matching.}
For each planar patch $\pi_j$, we encode the geometric information of nearby points into a compact descriptor named the Plane Context Histogram (PCH), illustrated in Fig.~\ref{fig:pch-construction}.
For each point $p_i$, we calculate its signed distance $\delta_i=u_j^\top(p_i-\mu_j)$ and its normal agreement $a_i=u_j^\top\nu_i$ with the patch.
We retain points satisfying $|\delta_i|\leq r_{\mathrm{PCH}}$, where $r_{\mathrm{PCH}}$ is a parameter controlling the signed-distance support of the descriptor, as shown in Fig.~\ref{fig:pch-construction}(a).
The resulting $(\delta_i,a_i)$ pairs are accumulated into a two-dimensional histogram with $16$ distance bins and $12$ normal-agreement bins, producing a $192$-dimensional descriptor as illustrated in Fig.~\ref{fig:pch-construction}(b).
The histogram is normalized to sum to one, reducing its sensitivity to the number of contributing points.

\begin{figure}[!t]
  \centering
  \begin{subfigure}[b]{0.48\columnwidth}
    \centering
    \includegraphics{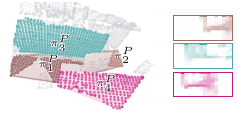}
    \caption{Plane-level descriptors in $P$}
    \label{fig:pch-matching-source}
  \end{subfigure}
  \hfill
  \begin{subfigure}[b]{0.48\columnwidth}
    \centering
    \includegraphics{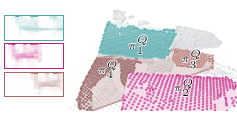}
    \caption{Plane-level descriptors in $Q$}
    \label{fig:pch-matching-target}
  \end{subfigure}

  \begin{subfigure}[b]{\columnwidth}
    \centering
    \includegraphics{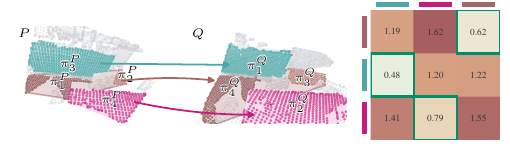}
    \caption{Plane-level matching}
    \label{fig:pch-matching-plane-level}
  \end{subfigure}

  \begin{subfigure}[b]{\columnwidth}
    \centering
    \includegraphics{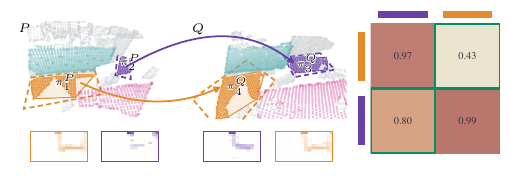}
    \caption{Coplanar-group refinement}
    \label{fig:pch-matching-refinement}
  \end{subfigure}

  \caption{
  Plane Context Histogram matching and coplanar-group refinement.
  (a,b) Individual patches are represented by their PCHs, while disconnected
  patches lying on the same plane are grouped and represented by a combined
  descriptor.
  (c) Plane-level descriptors are matched between $P$ and $Q$ using mutual
  nearest neighbors under the $\chi^2$-distance; the matrix shows the
  descriptor distances.
  (d) For a matched coplanar group, oriented bounding boxes restrict the
  support of each constituent patch to construct localized PCHs.
  These descriptors are then rematched within the group using the shown
  $2\times2$ distance matrix.
  }
  \label{fig:pch-matching}
\end{figure}

A point cloud may contain several disconnected patches that lie on the same infinite plane.
Because these patches produce very similar descriptors, we group them and combine their histograms into a single plane-level descriptor, as illustrated in Fig.~\ref{fig:pch-matching}(a,b).
The resulting plane-level descriptors are matched across the two point clouds using mutual nearest neighbors under the $\chi^2$-distance (Fig.~\ref{fig:pch-matching}(c)).

When a matched plane-level descriptor represents a single patch in each point cloud, the correspondence is obtained directly.
If it represents multiple disconnected coplanar patches, we perform a second matching stage within the matched groups, as illustrated in Fig.~\ref{fig:pch-matching}(d).
For each patch, we compute an oriented bounding box and construct a separate PCH using only points whose projections lie inside the box.
This captures differences in the local surroundings of each patch, helping distinguish patches that lie on the same plane.
We then apply the same matching procedure within each group.
The direct and refined matches together form the candidate plane-correspondence set $\mathcal C_{\mathrm{pl}}$.

For each correspondence, we also compute a confidence score by comparing the descriptor distance of the selected match with the descriptor distances to the next-best candidate for each patch.
For a mutual nearest-neighbor match, let $d=d_P^{(1)}=d_Q^{(1)}$ denote the descriptor distance between the matched patches, and let $d_P^{(2)}$ and $d_Q^{(2)}$ denote the descriptor distances to their respective second-best candidate matches.
We define the confidence score as
\begin{equation}
\label{eq:plane-match-confidence}
w =
\sqrt{
\left(1-\frac{d}{d_P^{(2)}}\right)
\left(1-\frac{d}{d_Q^{(2)}}\right)
}.
\end{equation}
Because $d\leq d_P^{(2)}$ and $d\leq d_Q^{(2)}$ for a mutual nearest-neighbor match, $w\in[0,1]$.
The score approaches one when the selected match is substantially closer than the next-best alternatives for both patches, and approaches zero when either descriptor is ambiguous.
This confidence score is later used to weight plane correspondences during graph-based outlier rejection.

\subsection{Outlier Rejection}
\label{sec:outlier-rejection}
\begin{figure}[t]
  \centering

  \begin{subfigure}[b]{\columnwidth}
    \centering
    \includegraphics{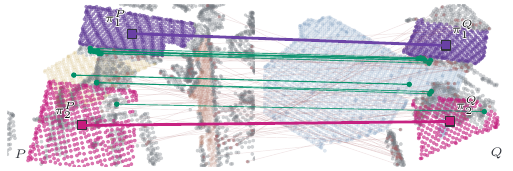}
    \caption{Candidate point and plane correspondences}
    \label{fig:outlier-rejection-candidates}
  \end{subfigure}

  \begin{subfigure}[b]{0.48\columnwidth}
    \centering
    \includegraphics{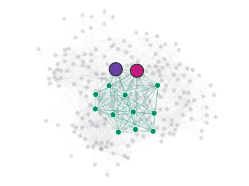}
    \caption{Maximum-weight joint clique}
    \label{fig:outlier-rejection-weighted}
  \end{subfigure}
  \hfill
  \begin{subfigure}[b]{0.48\columnwidth}
    \centering
    \includegraphics{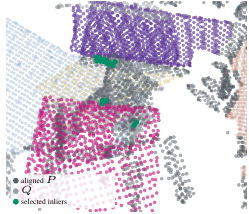}
    \caption{Estimated alignment}
    \label{fig:outlier-rejection-alignment}
  \end{subfigure}

  \begin{subfigure}[b]{0.48\columnwidth}
    \centering
    \includegraphics{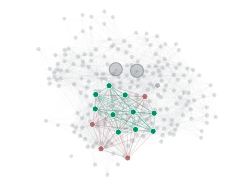}
    \caption{Maximum-cardinality point clique}
    \label{fig:outlier-rejection-cardinality}
  \end{subfigure}
  \hfill
  \begin{subfigure}[b]{0.48\columnwidth}
    \centering
    \includegraphics{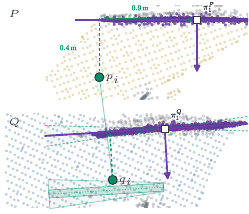}
    \caption{Adaptive plane--point bound}
    \label{fig:outlier-rejection-adaptive-bound}
  \end{subfigure}

\caption{Joint point--plane outlier rejection.
(a) Descriptor matching produces candidate point and plane correspondences.
(b) The maximum-weight clique combines the selected plane correspondences
(large colored nodes) with a mutually consistent subset of point correspondences (green nodes).
(c) The registration result from selected nodes.
(d) In contrast, selecting the clique by size yields a larger set that includes incorrect correspondences.
(e) The plane--point consistency threshold adapts to the plane-normal uncertainty and point-to-centroid distance.
}
  \label{fig:outlier-rejection}
\end{figure}

The candidate point and plane correspondences generally contain many incorrect matches.
We reject them by finding a subset of correspondences that are mutually consistent under a rigid transformation.
Let $\mathcal C_{\mathrm{pt}}=\{(p_i,q_i)\}$ and $\mathcal C_{\mathrm{pl}}=\{(\pi_j^P,\pi_j^Q)\}$ denote the candidate point and plane correspondences.
An example of the resulting point and plane correspondences is shown in Fig.~\ref{fig:outlier-rejection}(a).
Rigid transformations preserve several geometric relationships, which allow us to test the consistency of pairs of correspondences.

\textbf{Point--point consistency.}
For two inlier point correspondences $(p_i,q_i)$ and $(p_k,q_k)$, their pairwise distance is preserved:
\begin{equation}
\label{eq:point-point-consistency}
\left|
\|p_i-p_k\|_2-\|q_i-q_k\|_2
\right|
\leq 2\epsilon_p .
\end{equation}

\textbf{Plane--plane consistency.}
For two inlier plane correspondences $(\pi_j^P,\pi_j^Q)$ and $(\pi_k^P,\pi_k^Q)$, the angle between their normals is preserved:
\begin{equation}
\label{eq:plane-plane-consistency}
\left|
\angle(u_j^P,u_k^P)
-
\angle(u_j^Q,u_k^Q)
\right|
\leq 2\epsilon_\theta .
\end{equation}

\textbf{Plane--point consistency.}
A rigid transformation also preserves the signed distance between a point and
a plane.
For a plane correspondence $(\pi_j^P,\pi_j^Q)$ and point correspondence
$(p_i,q_i)$, we define
\begin{equation}
\label{eq:plane-point-residual}
r_{ij}
=
\left|
\left((u_j^P)^\top p_i-d_j^P\right)
-
\left((u_j^Q)^\top q_i-d_j^Q\right)
\right|.
\end{equation}
Normal uncertainty motivates the following bound:
\begin{equation}
\label{eq:plane-point-consistency}
r_{ij}
\leq
\epsilon_p
+\epsilon_\theta
\min\!\left(
\|p_i-\mu_j^P\|_2,
\|q_i-\mu_j^Q\|_2
\right),
\end{equation}
where $\epsilon_\theta$ is expressed in radians.
The first term accounts for the base distance error, while the second grows with the distance from the corresponding plane centroid.
Therefore, normal uncertainty has a larger effect on points farther from the plane centroid.
This effect is illustrated in Fig.~\ref{fig:outlier-rejection}(e), where uncertainty in the plane normal produces a larger bound farther from the plane centroid.

We construct a compatibility graph with one node for each candidate point or plane correspondence.
Two nodes are connected when their correspondences satisfy the applicable consistency test above. Therefore, every clique represents a set of pairwise consistent correspondences.

For each plane correspondence, we convert the descriptor confidence $w_j$ defined in \eqref{eq:plane-match-confidence} to the integer graph weight $\omega_j=\lceil 10w_j\rceil$.
Point correspondences are assigned a unit graph weight, while plane correspondences are assigned weight $\omega_j$.
We then select a maximum-weight clique rather than simply the clique with the largest number of nodes.
This favors sets containing distinctive plane matches without relaxing any of the geometric consistency tests.
Fig.~\ref{fig:outlier-rejection}(b) shows an example in which the plane correspondences help select the correct joint clique, while Fig.~\ref{fig:outlier-rejection}(d) shows the larger but incorrect point-only clique obtained when correspondences are selected only by cardinality.
A weighted clique can be represented in an unweighted graph by duplicating a node according to its integer weight.
Rather than explicitly expanding the graph in this way, we extend the Parallel Maximum Clique (PMC) solver to operate directly on nonnegative integer node weights.
This produces the same weighted-clique objective while avoiding the additional nodes and edges introduced by duplication.
The correspondences retained by the joint clique are then used for transformation estimation, producing the alignment shown in Fig.~\ref{fig:outlier-rejection}(c).

\subsection{Transformation Estimation}
After outlier rejection, let 
${\widehat{\mathcal C}_{\mathrm{pt}} =\{(p_i,q_i)\}_{i=1}^{N_{\mathrm{pt}}}}$
and
${\widehat{\mathcal C}_{\mathrm{pl}} =\{(\pi^P_j,\pi^Q_j)\}_{j=1}^{N_{\mathrm{pl}}}}$
denote the remaining correspondences.
We estimate the rigid transformation $(R,t)$ by solving
\begin{equation}
\label{eq:transformation-estimation}
\begin{aligned}
(R^\star,t^\star)
=
\underset{\substack{R\in\mathrm{SO}(3)\\t\in\mathbb{R}^3}}
{\arg\min}
\quad
&\sum_{i=1}^{N_{\mathrm{pt}}}
\left\|Rp_i+t-q_i\right\|_2^2
\\
&+
\sum_{j=1}^{N_{\mathrm{pl}}}
w_j\left\|Ru_j^P-u_j^Q\right\|_2^2
\\
&+
\sum_{j=1}^{N_{\mathrm{pl}}}
w_j
\left(
d_j^Q-d_j^P-(u_j^Q)^\top t
\right)^2 .
\end{aligned}
\end{equation}
The first term minimizes the Euclidean distance between corresponding points.
The second aligns the normals of corresponding planes.
The third aligns their offsets after applying the translation.
Each plane correspondence is weighted by its confidence score $w_j$.

The joint problem in \eqref{eq:transformation-estimation} does not admit a straightforward solution for $R$ and $t$ simultaneously.
However, when $t$ is fixed, optimizing $R$ reduces to an orthogonal Procrustes problem involving the matched points and plane normals, which can be solved using a $3\times3$ SVD.
Similarly, when $R$ is fixed, optimizing $t$ is a quadratic least-squares problem with a closed-form solution.
We therefore minimize the objective using Block Coordinate Descent (BCD), alternating between these two updates.
Since each update optimizes one variable while keeping the other fixed, the objective cannot increase between iterations.

To initialize BCD, we choose the initial translation so that the first rotation update reduces to an extended Arun solution \cite{arun1987least}.
Specifically, we set $t^{(0)}=\bar q$, where $\bar q$ is the centroid of the matched target points.
This reduces the point contribution of the first rotation update to the centered formulation of the classical Arun method, while the matched plane normals provide additional rotation constraints.
The resulting closed-form rotation provides the initialization for the subsequent BCD updates.

\begin{algorithm}[t]
\caption{Joint Point--Plane Transformation Estimation}
\label{alg:joint-point-plane-estimation}
\small
\begin{algorithmic}[1]

\State \textbf{Input}
Point correspondences
$\widehat{\mathcal C}_{\mathrm{pt}}
=\{(p_i,q_i)\}_{i=1}^{N_{\mathrm{pt}}}$,
plane correspondences
$\widehat{\mathcal C}_{\mathrm{pl}}
=\{(\pi_j^P,\pi_j^Q)\}_{j=1}^{N_{\mathrm{pl}}}$,
and plane weights $\{w_j\}$

\State \textbf{Output} Rigid transformation $(R,t)$

\Statex
\State \algcomment{Initialization}
\State
$t \gets \frac{1}{N_{\mathrm{pt}}}\sum_i q_i$
\hfill \algcomment{Extended Arun initialization}

\Statex
\State \algcomment{Block coordinate descent}
\Repeat

\State \algcomment{Update $R$ for fixed $t$}
\State
$H \gets
\sum_i (q_i-t)p_i^\top
+
\sum_j w_j u_j^Q (u_j^P)^\top$

\State
$H=U\Sigma V^\top$

\State
$R\gets
U\,\mathrm{diag}\!\left(1,1,\det(UV^\top)\right)V^\top$

\Statex
\State \algcomment{Update $t$ for fixed $R$}
\State
$A \gets
N_{\mathrm{pt}}I_3
+
\sum_j w_j u_j^Q(u_j^Q)^\top$

\State
$b \gets
\sum_i(q_i-Rp_i)
+
\sum_j w_j(d_j^Q-d_j^P)u_j^Q$

\State
Solve $At=b$ for $t$

\Until{$R$ and $t$ have converged}

\State \Return $(R,t)$

\end{algorithmic}
\end{algorithm}

\begin{figure}[!htbp]
  \centering

  \begin{subfigure}[b]{0.31\columnwidth}
    \centering
    \includegraphics[width=\linewidth]{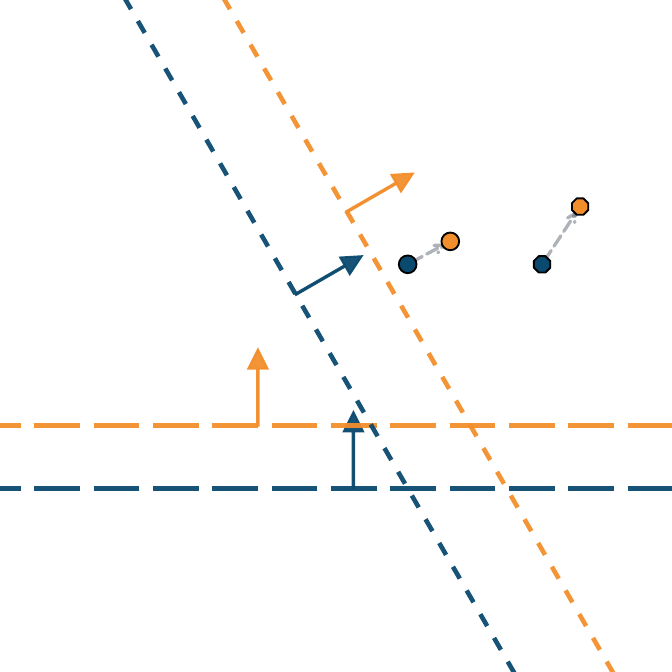}
    \caption{Initial pose}
    \label{fig:arun-initial}
  \end{subfigure}
  \hfill
  \begin{subfigure}[b]{0.31\columnwidth}
    \centering
    \includegraphics[width=\linewidth]{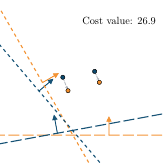}
    \caption{First BCD iteration}
    \label{fig:arun-earun}
  \end{subfigure}
  \hfill
  \begin{subfigure}[b]{0.31\columnwidth}
    \centering
    \includegraphics[width=\linewidth]{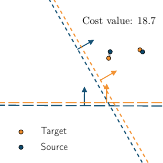}
    \caption{Final BCD result}
    \label{fig:arun-refinement}
  \end{subfigure}

  \caption{Comparison of alignment stages in BCD optimization, Algorithm~\ref{alg:joint-point-plane-estimation}.
  Source points and planes (orange) are progressively aligned to the target configuration (blue).
  (a) shows the initial mismatch.
  (b) shows the pose obtained after the first rotation update of the BCD scheme with fixed translation $t = q_c$, which is equivalent to the classical Arun rotation estimate.
  (c) shows the final result after the full nonlinear BCD optimization, where both point correspondences and planes are jointly refined.}
  \label{fig:arun-divergent-stages}
\end{figure}

\section{Experimental Evaluation}
In this section, we evaluate PARTE at three different levels.
First, we evaluate the proposed plane descriptor and matching procedure directly by measuring the quality of the resulting plane correspondences.
We then evaluate the complete registration pipeline across indoor RGB-D, outdoor LiDAR, and partial-to-full registration problems.
Finally, we study the main design choices through ablation experiments and analyze the runtime of the complete pipeline.

\subsection{Experimental Setup}
\textbf{Datasets.}
We organize the evaluation around established registration benchmarks and follow their standard pair definitions and success criteria whenever available.
For indoor RGB-D registration, we use the eight test scenes of 3DMatch~\cite{zeng20173dmatch} and additionally report results on 3DLoMatch~\cite{huang2021predator}.
3DMatch contains pairs with at least $30\%$ overlap, while 3DLoMatch contains the more challenging pairs with $10$--$30\%$ overlap.
For outdoor LiDAR registration, we use the KITTI odometry dataset~\cite{Geiger2012CVPR} under two complementary protocols.
The first is the standard KITTI-10m pairwise-registration benchmark commonly used in prior work~\cite{bai2020d3feat,huang2021predator}, while the second is the KITTI-LC loop-closure benchmark introduced by G3Reg~\cite{qiao2024g3reg}.
We further include two additional settings that test registration outside these standard benchmarks.
From RESSO~\cite{chen2019plade}, we select sequences in which a local hand-held scan is registered against a larger static scan, forming a challenging partial-to-full registration problem.
Finally, we use the ETH benchmark~\cite{pomerleau2012challenging} to evaluate full-scan registration on challenging terrestrial LiDAR data.

\begin{table}[!t]
\centering
\scriptsize
\setlength{\tabcolsep}{2pt}
\resizebox{\columnwidth}{!}{%
\begin{tabular}{@{}l c c c c c c l@{}}
\toprule
Dataset & Sensor & \#Scenes & \#Pairs & $v$ & \#Pts/scan & \#Planes/scan & Success criterion \\[2pt]
\midrule
3DMatch~\cite{zeng20173dmatch} & RGB-D & 8 & 1,623 & 5 cm & 4,838 & 4.6 & $\leq 15^{\circ}$, $\leq 30$ cm \\
3DLoMatch~\cite{huang2021predator} & RGB-D & 8 & 1,781 & 5 cm & 4,821 & 4.6 & $\leq 15^{\circ}$, $\leq 30$ cm \\
\midrule
KITTI-10m~\cite{qiao2024g3reg} & LiDAR & 3 & 556 & 30 cm & 18,238 & 5.1 & $\leq 5^{\circ}$, $\leq 0.6$ m \\
KITTI-LC~\cite{qiao2024g3reg} & LiDAR & 5 & 3,325 & 30 cm & 18,354 & 7.9 & $\leq 5^{\circ}$, $\leq 2$ m \\
\midrule
RESSO P2F~\cite{chen2019plade} & TLS + RGB-D & 7 & 99 & 5 cm & 10,143 & 7.4 & $\leq 15^{\circ}$, $\leq 30$ cm \\
ETH~\cite{pomerleau2012challenging} & LiDAR & 4 & 713 & 10 cm & 30,234 & 2.0 & $\leq 15^{\circ}$, $\leq 30$ cm \\
\bottomrule
\end{tabular}%

}
\caption{Datasets and evaluation protocols used in our experiments.
Point and plane counts are averages over unique scans appearing in the
evaluation pairs after voxel downsampling.}
\label{tab:datasets}
\end{table}

\textbf{Compared methods.}
For the plane-matching evaluation, we compare PCH against a broad set of
descriptors that represent planar patches using different geometric cues.
As simple geometric baselines, we use patch area and perimeter.
The local descriptor baselines include Spin Images~\cite{johnson1997spin},
Angular Spin Images~\cite{endres2009unsupervised}, USC~\cite{tombari2010unique}, and 3D Shape Context~\cite{kortgen20033d}.
We additionally evaluate global and segment-level representations, including VFH~\cite{rusu2010vfh}, CVFH~\cite{aldoma2011cad}, OUR-CVFH~\cite{aldoma2012ourcvfh},
GRSD~\cite{marton2011combined,kanezaki2011voxelized},
ESF~\cite{wohlkinger2011esf}, GASD~\cite{lima2016gasd}, the handcrafted segment features used by SegMatch~\cite{dube2017segmatch},
and the Gaussian Ellipsoid Model (GEM) representation used by
G3Reg~\cite{qiao2024g3reg}.

For the complete registration experiments, we compare PARTE with representative classical, graph-based, geometry-based, and learning-based methods.
The classical baselines include RANSAC~\cite{fischler1981random} and FGR~\cite{zhou2016fast}.
We compare against robust correspondence-filtering methods TEASER++~\cite{yang2020teaser}, MAC~\cite{zhang20233d}, and CLIPPER+~\cite{fathian2024clipper+}, as well as the QUATRO estimator~\cite{lim2022quatro}, evaluated with ground-filtered point correspondences following the QUATRO++ strategy~\cite{lim2024quatropp}.
G3Reg~\cite{qiao2024g3reg} provides a recent geometry-based comparison using higher-level geometric primitives.
We additionally compare against the plane-based methods PLADE~\cite{chen2019plade} and VPFBR~\cite{li2022point}.
For datasets with publicly available trained models, we additionally compare against the learning-based methods PREDATOR~\cite{huang2021predator}, PointDSC~\cite{bai2021pointdsc}, and VBReg~\cite{jiang2023robust}.

\textbf{Evaluation metrics.}
For plane matching, we report F1 score, average precision (AP), and the average
number of true-positive correspondences per scan pair, $\overline{\mathrm{TP}}$.
F1 is computed from the mutual nearest-neighbor correspondences, while AP
summarizes the precision--recall curve obtained by varying the descriptor
distance threshold over all candidate plane pairs.
For registration, given an estimated transformation $(R,t)$ and ground truth
$(R^\star,t^\star)$, we compute the rotation error (RE)
\[
\mathrm{RE}
=
\arccos\!\left(
\frac{\operatorname{tr}(R^\top R^\star)-1}{2}
\right),
\]
and translation error (TE)
\[
\mathrm{TE}=\lVert t-t^\star\rVert_2.
\]
We report the success rate (SR), defined as the percentage of registration pairs whose RE and TE satisfy the dataset-specific criteria in Table~\ref{tab:datasets}.
We additionally report average rotation error (ARE), average translation error (ATE), and mean end-to-end runtime per pair.
Unless otherwise stated, RE and ARE are reported in degrees, TE and ATE in centimeters, and runtime in milliseconds.

\textbf{Implementation and parameter settings.}
Table~\ref{tab:parameters} summarizes the parameters used throughout the experiments.
We use the voxel size $v$ as the common geometric scale of the pipeline.
Apart from selecting $v$ for each dataset as shown in Table~\ref{tab:datasets}, the remaining parameters are fixed across all experiments or expressed directly as multiples of $v$.
Surface normals are estimated within a radius of $2v$ using at most 30 neighbors, and the same neighborhoods are used during plane extraction.
Plane segmentation uses $\tau_d=v$, $\tau_\theta=0.2$, and a minimum patch size of 100 points.
FPFH descriptors use a radius of $5v$ with at most 100 neighbors, while PCH uses $r_{\mathrm{PCH}}=20v$ and therefore covers signed distances in the range $[-20v,20v]$.
For outlier rejection, we use the base spatial tolerance $\epsilon_p=v$ and the base angular tolerance $\epsilon_\theta=5^\circ$.
The resulting point--point and plane--plane bounds are therefore $2v$ and $10^\circ$, respectively, while the plane--point bound adapts to the matched-plane geometry as described in Sec.~\ref{sec:outlier-rejection}.

\begin{table}[t]
\centering
\caption{Parameter settings used throughout the experiments.}
\label{tab:parameters}
\scriptsize
\setlength{\tabcolsep}{3pt}

\begin{tabularx}{\columnwidth}{@{}llXl@{}}
\toprule
Stage & Symbol & Parameter & Value \\
\midrule

Normal estimation
    & -- & Radius / max neighbors & $2v$ / 30 \\

\addlinespace[1pt]
Plane extraction
    & $\tau_d$ & Thickness & $v$ \\
    & $\tau_\theta$ & Normal dispersion & $0.2$ \\
    & -- & Minimum points & 100 \\

\addlinespace[1pt]
FPFH
    & -- & Radius / max neighbors & $5v$ / 100 \\

\addlinespace[1pt]
PCH
    & $r_{\mathrm{PCH}}$ & Signed-distance half-range & $20v$ \\

\addlinespace[1pt]
Outlier rejection
    & $\epsilon_p$ & Base spatial tolerance & $v$ \\
    & $\epsilon_\theta$ & Base angular tolerance & $5^\circ$ \\

\bottomrule
\end{tabularx}
\end{table}

\subsection{Plane Descriptor Evaluation}
\label{sec:plane-descriptor-evaluation}

\begin{table*}[!t]
\centering
\setlength{\tabcolsep}{2pt}
\renewcommand{\arraystretch}{0.9}
\resizebox{\textwidth}{!}{%
\begin{tabular}{llc@{\hspace{4pt}}ccc!{\color{white}\vrule width 5pt}ccc!{\color{white}\vrule width 5pt}ccc!{\color{white}\vrule width 5pt}ccc!{\color{white}\vrule width 5pt}ccc!{\color{white}\vrule width 5pt}ccc}
\toprule
\multirow{2}{*}{Name} & \multirow{2}{*}{Dist.} & \multirow{2}{*}{Size} & \multicolumn{3}{c}{3DLoMatch} & \multicolumn{3}{c}{3DMatch} & \multicolumn{3}{c}{KITTI-10m} & \multicolumn{3}{c}{KITTI-LC} & \multicolumn{3}{c}{ETH} & \multicolumn{3}{c}{RESSO} \\
 &  &  & F1 & AP & $\overline{\mathrm{TP}}$ & F1 & AP & $\overline{\mathrm{TP}}$ & F1 & AP & $\overline{\mathrm{TP}}$ & F1 & AP & $\overline{\mathrm{TP}}$ & F1 & AP & $\overline{\mathrm{TP}}$ & F1 & AP & $\overline{\mathrm{TP}}$ \\
\midrule
\multicolumn{21}{l}{\textit{Simple geometry}} \\
Perimeter & $|\log(a/b)|$ & 1 & \cellcolor{ResultRed!64!ResultYellow}0.15 & \cellcolor{ResultRed!81!ResultYellow}0.08 & \cellcolor{ResultRed!57!ResultYellow}0.26 & \cellcolor{ResultRed!21!ResultYellow}0.31 & \cellcolor{ResultRed!50!ResultYellow}0.21 & \cellcolor{ResultRed!38!ResultYellow}0.83 & \cellcolor{ResultRed!61!ResultYellow}0.16 & \cellcolor{ResultRed!79!ResultYellow}0.08 & \cellcolor{ResultRed!69!ResultYellow}0.78 & \cellcolor{ResultRed!75!ResultYellow}0.10 & \cellcolor{ResultRed!88!ResultYellow}0.05 & \cellcolor{ResultRed!79!ResultYellow}0.69 & \cellcolor{ResultYellow!36!ResultGreen}0.65 & \cellcolor{ResultYellow!84!ResultGreen}0.48 & \cellcolor{ResultYellow!37!ResultGreen}0.90 & \cellcolor{ResultRed!86!ResultYellow}0.05 & \cellcolor{ResultRed!94!ResultYellow}0.02 & \cellcolor{ResultRed!87!ResultYellow}0.21 \\
Area & $|\log(a/b)|$ & 1 & \cellcolor{ResultRed!62!ResultYellow}0.15 & \cellcolor{ResultRed!80!ResultYellow}0.08 & \cellcolor{ResultRed!54!ResultYellow}0.27 & \cellcolor{ResultRed!21!ResultYellow}0.32 & \cellcolor{ResultRed!49!ResultYellow}0.21 & \cellcolor{ResultRed!37!ResultYellow}0.84 & \cellcolor{ResultRed!66!ResultYellow}0.13 & \cellcolor{ResultRed!81!ResultYellow}0.08 & \cellcolor{ResultRed!74!ResultYellow}0.66 & \cellcolor{ResultRed!79!ResultYellow}0.08 & \cellcolor{ResultRed!89!ResultYellow}0.05 & \cellcolor{ResultRed!82!ResultYellow}0.59 & \cellcolor{ResultYellow!35!ResultGreen}0.66 & \cellcolor{ResultYellow!75!ResultGreen}0.51 & \cellcolor{ResultYellow!37!ResultGreen}0.90 & \cellcolor{ResultRed!88!ResultYellow}0.05 & \cellcolor{ResultRed!94!ResultYellow}0.02 & \cellcolor{ResultRed!89!ResultYellow}0.19 \\
\addlinespace[2pt]
\multicolumn{21}{l}{\textit{Local point descriptors}} \\
Angular Spin~\cite{endres2009unsupervised} & $1-\rho$ & 153 & \cellcolor{ResultRed!68!ResultYellow}0.13 & \cellcolor{ResultRed!80!ResultYellow}0.08 & \cellcolor{ResultRed!68!ResultYellow}0.19 & \cellcolor{ResultRed!41!ResultYellow}0.24 & \cellcolor{ResultRed!53!ResultYellow}0.19 & \cellcolor{ResultRed!59!ResultYellow}0.54 & \cellcolor{ResultRed!70!ResultYellow}0.12 & \cellcolor{ResultRed!79!ResultYellow}0.09 & \cellcolor{ResultRed!80!ResultYellow}0.50 & \cellcolor{ResultRed!82!ResultYellow}0.07 & \cellcolor{ResultRed!89!ResultYellow}0.05 & \cellcolor{ResultRed!88!ResultYellow}0.42 & \cellcolor{ResultRed!37!ResultYellow}0.25 & \cellcolor{ResultRed!65!ResultYellow}0.14 & \cellcolor{ResultRed!60!ResultYellow}0.22 & \cellcolor{ResultRed!85!ResultYellow}0.06 & \cellcolor{ResultRed!92!ResultYellow}0.03 & \cellcolor{ResultRed!88!ResultYellow}0.20 \\
Spin Image~\cite{johnson1997spin} & $1-\rho$ & 153 & \cellcolor{ResultRed!70!ResultYellow}0.12 & \cellcolor{ResultRed!79!ResultYellow}0.09 & \cellcolor{ResultRed!69!ResultYellow}0.19 & \cellcolor{ResultRed!32!ResultYellow}0.27 & \cellcolor{ResultRed!44!ResultYellow}0.23 & \cellcolor{ResultRed!52!ResultYellow}0.64 & \cellcolor{ResultRed!57!ResultYellow}0.17 & \cellcolor{ResultRed!71!ResultYellow}0.12 & \cellcolor{ResultRed!70!ResultYellow}0.75 & \cellcolor{ResultRed!73!ResultYellow}0.11 & \cellcolor{ResultRed!84!ResultYellow}0.07 & \cellcolor{ResultRed!80!ResultYellow}0.66 & \cellcolor{ResultYellow!27!ResultGreen}0.69 & \cellcolor{ResultYellow!35!ResultGreen}0.68 & \cellcolor{ResultYellow!36!ResultGreen}0.91 & \cellcolor{ResultRed!80!ResultYellow}0.08 & \cellcolor{ResultRed!92!ResultYellow}0.03 & \cellcolor{ResultRed!84!ResultYellow}0.27 \\
USC~\cite{tombari2010unique} & $\ell_2$ & 1960 & \cellcolor{ResultRed!58!ResultYellow}0.17 & \cellcolor{ResultRed!75!ResultYellow}0.10 & \cellcolor{ResultRed!68!ResultYellow}0.19 & \cellcolor{ResultRed!27!ResultYellow}0.29 & \cellcolor{ResultRed!35!ResultYellow}0.27 & \cellcolor{ResultRed!58!ResultYellow}0.55 & \cellcolor{ResultRed!65!ResultYellow}0.14 & \cellcolor{ResultRed!67!ResultYellow}0.13 & \cellcolor{ResultRed!83!ResultYellow}0.42 & \cellcolor{ResultRed!77!ResultYellow}0.09 & \cellcolor{ResultRed!82!ResultYellow}0.08 & \cellcolor{ResultRed!89!ResultYellow}0.36 & \cellcolor{ResultYellow!40!ResultGreen}0.64 & \cellcolor{ResultYellow!54!ResultGreen}0.60 & \cellcolor{ResultYellow!77!ResultGreen}0.68 & \cellcolor{ResultRed!84!ResultYellow}0.06 & \cellcolor{ResultRed!90!ResultYellow}0.04 & \cellcolor{ResultRed!91!ResultYellow}0.15 \\
3DSC~\cite{kortgen20033d} & $\chi^2$ & 1980 & \cellcolor{ResultRed!62!ResultYellow}0.15 & \cellcolor{ResultRed!71!ResultYellow}0.12 & \cellcolor{ResultRed!68!ResultYellow}0.19 & \cellcolor{ResultRed!10!ResultYellow}0.36 & \cellcolor{ResultRed!24!ResultYellow}0.32 & \cellcolor{ResultRed!44!ResultYellow}0.74 & \cellcolor{ResultRed!37!ResultYellow}0.25 & \cellcolor{ResultRed!55!ResultYellow}0.19 & \cellcolor{ResultRed!60!ResultYellow}0.99 & \cellcolor{ResultRed!59!ResultYellow}0.16 & \cellcolor{ResultRed!74!ResultYellow}0.11 & \cellcolor{ResultRed!74!ResultYellow}0.86 & \cellcolor{ResultYellow!20!ResultGreen}0.72 & \cellcolor{ResultYellow!12!ResultGreen}0.77 & \cellcolor{ResultYellow!36!ResultGreen}0.91 & \cellcolor{ResultRed!87!ResultYellow}0.05 & \cellcolor{ResultRed!91!ResultYellow}0.04 & \cellcolor{ResultRed!92!ResultYellow}0.14 \\
\addlinespace[2pt]
\multicolumn{21}{l}{\textit{Global/segment descriptors}} \\
OUR-CVFH~\cite{aldoma2012ourcvfh} & min $\ell_1$ & $K \times 308$ & \cellcolor{ResultRed!46!ResultYellow}0.22 & \cellcolor{ResultRed!65!ResultYellow}0.14 & \cellcolor{ResultRed!39!ResultYellow}0.36 & \cellcolor{ResultYellow!83!ResultGreen}0.47 & \cellcolor{ResultYellow!97!ResultGreen}0.42 & \cellcolor{ResultRed!9!ResultYellow}1.21 & \cellcolor{ResultRed!35!ResultYellow}0.26 & \cellcolor{ResultRed!52!ResultYellow}0.20 & \cellcolor{ResultRed!51!ResultYellow}1.23 & \cellcolor{ResultRed!57!ResultYellow}0.17 & \cellcolor{ResultRed!71!ResultYellow}0.12 & \cellcolor{ResultRed!66!ResultYellow}1.13 & \cellcolor{ResultYellow!84!ResultGreen}0.46 & \cellcolor{ResultRed!16!ResultYellow}0.35 & \cellcolor{ResultRed!28!ResultYellow}0.40 & \cellcolor{ResultRed!69!ResultYellow}0.12 & \cellcolor{ResultRed!89!ResultYellow}0.05 & \cellcolor{ResultRed!75!ResultYellow}0.41 \\
CVFH~\cite{aldoma2011cad} & min $\ell_1$ & $K \times 308$ & \cellcolor{ResultRed!46!ResultYellow}0.22 & \cellcolor{ResultRed!64!ResultYellow}0.15 & \cellcolor{ResultRed!39!ResultYellow}0.36 & \cellcolor{ResultYellow!82!ResultGreen}0.47 & \cellcolor{ResultYellow!98!ResultGreen}0.42 & \cellcolor{ResultRed!9!ResultYellow}1.21 & \cellcolor{ResultRed!35!ResultYellow}0.26 & \cellcolor{ResultRed!53!ResultYellow}0.19 & \cellcolor{ResultRed!51!ResultYellow}1.22 & \cellcolor{ResultRed!59!ResultYellow}0.16 & \cellcolor{ResultRed!71!ResultYellow}0.12 & \cellcolor{ResultRed!68!ResultYellow}1.08 & \cellcolor{ResultYellow!8!ResultGreen}\underline{0.76} & \cellcolor{ResultYellow!1!ResultGreen}\underline{0.82} & \cellcolor{ResultYellow!16!ResultGreen}1.02 & \cellcolor{ResultRed!67!ResultYellow}0.13 & \cellcolor{ResultRed!85!ResultYellow}0.06 & \cellcolor{ResultRed!71!ResultYellow}0.48 \\
GRSD~\cite{marton2011combined,kanezaki2011voxelized} & $\chi^2$ & 21 & \cellcolor{ResultRed!77!ResultYellow}0.09 & \cellcolor{ResultRed!84!ResultYellow}0.07 & \cellcolor{ResultRed!75!ResultYellow}0.15 & \cellcolor{ResultRed!60!ResultYellow}0.16 & \cellcolor{ResultRed!66!ResultYellow}0.14 & \cellcolor{ResultRed!70!ResultYellow}0.39 & \cellcolor{ResultRed!78!ResultYellow}0.09 & \cellcolor{ResultRed!85!ResultYellow}0.06 & \cellcolor{ResultRed!83!ResultYellow}0.41 & \cellcolor{ResultRed!87!ResultYellow}0.05 & \cellcolor{ResultRed!91!ResultYellow}0.04 & \cellcolor{ResultRed!90!ResultYellow}0.34 & \cellcolor{ResultYellow!56!ResultGreen}0.57 & \cellcolor{ResultRed!3!ResultYellow}0.40 & \cellcolor{ResultYellow!70!ResultGreen}0.72 & \cellcolor{ResultRed!92!ResultYellow}0.03 & \cellcolor{ResultRed!95!ResultYellow}0.02 & \cellcolor{ResultRed!93!ResultYellow}0.11 \\
VFH~\cite{rusu2010vfh} & $\ell_1$ & 308 & \cellcolor{ResultRed!45!ResultYellow}0.22 & \cellcolor{ResultRed!65!ResultYellow}0.14 & \cellcolor{ResultRed!37!ResultYellow}0.38 & \cellcolor{ResultYellow!84!ResultGreen}0.46 & \cellcolor{ResultRed!1!ResultYellow}0.41 & \cellcolor{ResultRed!9!ResultYellow}1.20 & \cellcolor{ResultRed!42!ResultYellow}0.23 & \cellcolor{ResultRed!59!ResultYellow}0.17 & \cellcolor{ResultRed!55!ResultYellow}1.12 & \cellcolor{ResultRed!62!ResultYellow}0.15 & \cellcolor{ResultRed!76!ResultYellow}0.10 & \cellcolor{ResultRed!70!ResultYellow}1.02 & \cellcolor{ResultYellow!7!ResultGreen}\textbf{0.77} & \cellcolor{ResultYellow!0!ResultGreen}\textbf{0.82} & \cellcolor{ResultYellow!15!ResultGreen}1.02 & \cellcolor{ResultRed!72!ResultYellow}0.11 & \cellcolor{ResultRed!88!ResultYellow}0.05 & \cellcolor{ResultRed!75!ResultYellow}0.42 \\
GASD~\cite{lima2016gasd} & $\ell_1$ & 512 & \cellcolor{ResultRed!71!ResultYellow}0.11 & \cellcolor{ResultRed!82!ResultYellow}0.08 & \cellcolor{ResultRed!70!ResultYellow}0.18 & \cellcolor{ResultRed!27!ResultYellow}0.29 & \cellcolor{ResultRed!41!ResultYellow}0.24 & \cellcolor{ResultRed!48!ResultYellow}0.69 & \cellcolor{ResultRed!51!ResultYellow}0.20 & \cellcolor{ResultRed!68!ResultYellow}0.13 & \cellcolor{ResultRed!67!ResultYellow}0.81 & \cellcolor{ResultRed!68!ResultYellow}0.13 & \cellcolor{ResultRed!81!ResultYellow}0.08 & \cellcolor{ResultRed!79!ResultYellow}0.71 & \cellcolor{ResultYellow!29!ResultGreen}0.68 & \cellcolor{ResultYellow!42!ResultGreen}0.65 & \cellcolor{ResultYellow!37!ResultGreen}0.90 & \cellcolor{ResultRed!76!ResultYellow}0.09 & \cellcolor{ResultRed!91!ResultYellow}0.04 & \cellcolor{ResultRed!81!ResultYellow}0.31 \\
ESF~\cite{wohlkinger2011esf} & $\chi^2$ & 640 & \cellcolor{ResultRed!61!ResultYellow}0.15 & \cellcolor{ResultRed!75!ResultYellow}0.10 & \cellcolor{ResultRed!57!ResultYellow}0.26 & \cellcolor{ResultRed!19!ResultYellow}0.32 & \cellcolor{ResultRed!31!ResultYellow}0.29 & \cellcolor{ResultRed!38!ResultYellow}0.82 & \cellcolor{ResultRed!47!ResultYellow}0.21 & \cellcolor{ResultRed!64!ResultYellow}0.15 & \cellcolor{ResultRed!59!ResultYellow}1.01 & \cellcolor{ResultRed!63!ResultYellow}0.15 & \cellcolor{ResultRed!76!ResultYellow}0.10 & \cellcolor{ResultRed!71!ResultYellow}0.98 & \cellcolor{ResultYellow!38!ResultGreen}0.64 & \cellcolor{ResultYellow!47!ResultGreen}0.63 & \cellcolor{ResultYellow!52!ResultGreen}0.82 & \cellcolor{ResultRed!76!ResultYellow}0.10 & \cellcolor{ResultRed!89!ResultYellow}0.04 & \cellcolor{ResultRed!79!ResultYellow}0.35 \\
SegMatch~\cite{dube2017segmatch} & $\ell_2$ & 8 & \cellcolor{ResultRed!63!ResultYellow}0.15 & \cellcolor{ResultRed!79!ResultYellow}0.09 & \cellcolor{ResultRed!59!ResultYellow}0.24 & \cellcolor{ResultRed!18!ResultYellow}0.32 & \cellcolor{ResultRed!41!ResultYellow}0.24 & \cellcolor{ResultRed!37!ResultYellow}0.83 & \cellcolor{ResultRed!45!ResultYellow}0.22 & \cellcolor{ResultRed!67!ResultYellow}0.14 & \cellcolor{ResultRed!58!ResultYellow}1.05 & \cellcolor{ResultRed!64!ResultYellow}0.14 & \cellcolor{ResultRed!82!ResultYellow}0.07 & \cellcolor{ResultRed!71!ResultYellow}0.97 & \cellcolor{ResultYellow!33!ResultGreen}0.66 & \cellcolor{ResultYellow!63!ResultGreen}0.56 & \cellcolor{ResultYellow!49!ResultGreen}0.83 & \cellcolor{ResultRed!80!ResultYellow}0.08 & \cellcolor{ResultRed!89!ResultYellow}0.04 & \cellcolor{ResultRed!83!ResultYellow}0.28 \\
GEM~\cite{qiao2024g3reg} & $W_2^2$ & 12 & \cellcolor{ResultRed!59!ResultYellow}0.16 & \cellcolor{ResultRed!81!ResultYellow}0.08 & \cellcolor{ResultRed!53!ResultYellow}0.28 & \cellcolor{ResultRed!13!ResultYellow}0.35 & \cellcolor{ResultRed!45!ResultYellow}0.23 & \cellcolor{ResultRed!32!ResultYellow}0.89 & \cellcolor{ResultRed!43!ResultYellow}0.23 & \cellcolor{ResultRed!72!ResultYellow}0.12 & \cellcolor{ResultRed!55!ResultYellow}1.11 & \cellcolor{ResultRed!62!ResultYellow}0.15 & \cellcolor{ResultRed!85!ResultYellow}0.06 & \cellcolor{ResultRed!69!ResultYellow}1.05 & \cellcolor{ResultYellow!21!ResultGreen}0.71 & \cellcolor{ResultYellow!92!ResultGreen}0.44 & \cellcolor{ResultYellow!25!ResultGreen}0.96 & \cellcolor{ResultRed!76!ResultYellow}0.09 & \cellcolor{ResultRed!88!ResultYellow}0.05 & \cellcolor{ResultRed!79!ResultYellow}0.35 \\
\addlinespace[2pt]
\multicolumn{21}{l}{\textit{Ours}} \\
\textbf{PCH} & $\chi^2$ & 192 & \cellcolor{ResultYellow!92!ResultGreen}\underline{0.43} & \cellcolor{ResultRed!36!ResultYellow}\underline{0.26} & \cellcolor{ResultYellow!76!ResultGreen}\underline{0.74} & \cellcolor{ResultYellow!2!ResultGreen}\underline{0.79} & \cellcolor{ResultYellow!24!ResultGreen}\underline{0.72} & \cellcolor{ResultYellow!31!ResultGreen}\underline{2.23} & \cellcolor{ResultYellow!51!ResultGreen}\underline{0.59} & \cellcolor{ResultYellow!93!ResultGreen}\underline{0.44} & \cellcolor{ResultYellow!83!ResultGreen}\underline{2.91} & \cellcolor{ResultYellow!94!ResultGreen}\underline{0.42} & \cellcolor{ResultRed!28!ResultYellow}\underline{0.30} & \cellcolor{ResultRed!18!ResultYellow}\underline{2.76} & \cellcolor{ResultYellow!8!ResultGreen}0.76 & \cellcolor{ResultYellow!40!ResultGreen}0.66 & \cellcolor{ResultYellow!11!ResultGreen}\textbf{1.04} & \cellcolor{ResultRed!23!ResultYellow}\underline{0.31} & \cellcolor{ResultRed!55!ResultYellow}\textbf{0.18} & \cellcolor{ResultRed!37!ResultYellow}\underline{1.07} \\
\textbf{PCH-2L} & $\chi^2$ & 192 & \cellcolor{ResultYellow!85!ResultGreen}\textbf{0.46} & \cellcolor{ResultRed!17!ResultYellow}\textbf{0.34} & \cellcolor{ResultYellow!70!ResultGreen}\textbf{0.77} & \cellcolor{ResultYellow!0!ResultGreen}\textbf{0.80} & \cellcolor{ResultYellow!13!ResultGreen}\textbf{0.77} & \cellcolor{ResultYellow!30!ResultGreen}\textbf{2.24} & \cellcolor{ResultYellow!35!ResultGreen}\textbf{0.66} & \cellcolor{ResultYellow!68!ResultGreen}\textbf{0.54} & \cellcolor{ResultYellow!73!ResultGreen}\textbf{3.17} & \cellcolor{ResultYellow!80!ResultGreen}\textbf{0.48} & \cellcolor{ResultRed!19!ResultYellow}\textbf{0.33} & \cellcolor{ResultRed!11!ResultYellow}\textbf{3.00} & \cellcolor{ResultYellow!9!ResultGreen}0.76 & \cellcolor{ResultYellow!26!ResultGreen}0.72 & \cellcolor{ResultYellow!11!ResultGreen}\underline{1.04} & \cellcolor{ResultRed!19!ResultYellow}\textbf{0.32} & \cellcolor{ResultRed!56!ResultYellow}\underline{0.18} & \cellcolor{ResultRed!35!ResultYellow}\textbf{1.09} \\
\bottomrule
\end{tabular}%

}
\caption{Plane-matching performance across the evaluated datasets. We report F1 score, average precision (AP), and the average number of true-positive correspondences per scan pair, $\overline{\mathrm{TP}}$.}
\label{tab:plane-matching-results}
\end{table*}

We first evaluate the quality of the plane correspondences produced by PCH independently of the complete registration pipeline.
We compare against a broad set of alternatives, including simple geometric properties, local point descriptors, and global and segment descriptors.
All competing descriptors are computed from the same extracted planar patches. Patch-based descriptors use only the points belonging to the corresponding plane, whereas PCH additionally uses the surrounding scene geometry by design.
For each descriptor, we construct the pairwise distance matrix between the planes extracted from the two scans and establish correspondences using mutual nearest-neighbor (MNN) matching.
Different distance measures have been used with histogram-based geometric descriptors in prior work and reference implementations~\cite{rusu2010vfh,wohlkinger2011esf,kortgen20033d,aldoma2011cad,aldoma2012ourcvfh}.
We therefore evaluate multiple reasonable choices where applicable and report only the distance that gives the highest overall MNN F1 score across the evaluated datasets.
Specifically, we compare $\ell_1$ and $\chi^2$ distances for PCH, ESF, and VFH; minimum $\ell_1$ and minimum $\chi^2$ distances between the signature sets produced by CVFH and OUR-CVFH; and $\ell_2$ and normalized $\chi^2$ distances for 3DSC.
For Spin Image and Angular Spin, we use $1-\rho$, where $\rho$ denotes the Pearson correlation coefficient between the two descriptor vectors.
For the simple geometric baselines, we similarly evaluate absolute difference, ratio, and log-ratio measures.
The selected distance for each descriptor is reported in Table~\ref{tab:plane-matching-results}.

Table~\ref{tab:plane-matching-results} summarizes the resulting plane-matching performance.
F1 and $\overline{\mathrm{TP}}$ are computed from the MNN correspondences, while AP summarizes the precision--recall behavior of each descriptor.
\textbf{Our one-level PCH matcher performs substantially better than the competing descriptors} across most datasets, while \textbf{our full two-level matcher, PCH-2L} in Table~\ref{tab:plane-matching-results}, \textbf{further improves performance}.
In particular, the improvement is large on 3DLoMatch and the two KITTI benchmarks, where descriptors based primarily on the shape or extent of the planar patch provide relatively weak discrimination.

\begin{figure}[th]
        \centering
        \includegraphics[width=\columnwidth]{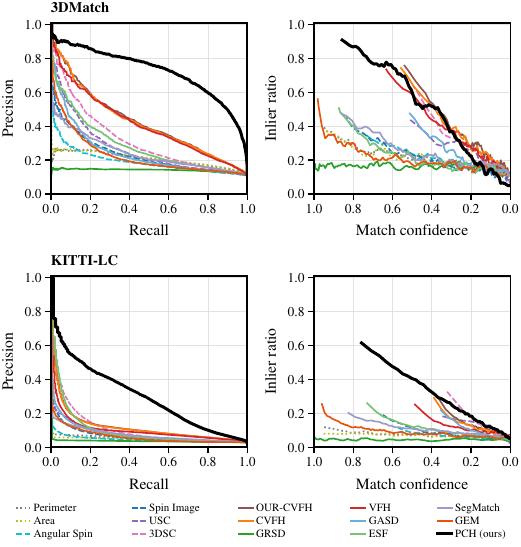}
        
        \caption{Plane matching performance on 3DMatch (top) and KITTI-LC (bottom). Left column: precision--recall curves, where PCH
        maintains substantially higher precision across the recall range.
        Right column: descriptor-match confidence versus empirical inlier
        ratio, where higher PCH confidence corresponds strongly to a higher
        probability of a correct plane correspondence.}
        \label{fig:plane-matching-combined}
\end{figure}

Fig.~\ref{fig:plane-matching-combined} shows the corresponding precision--recall curves and confidence analysis on 3DMatch and KITTI-LC.
\textbf{PCH remains above the competing precision--recall curves throughout the plotted recall range on both datasets}, showing that its advantage is not limited to a particular matching threshold or method.
What sets PCH apart is that it describes each plane using the geometry around it rather than the planar patch alone.
Overall, these results support the main motivation for PCH: planar patches are often difficult to distinguish from their own geometry alone, while the geometry surrounding the plane provides a more distinctive matching cue.

Fig.~\ref{fig:plane-matching-combined} also shows how the empirical inlier ratio varies with the match confidence defined in \eqref{eq:plane-match-confidence}.
For PCH, the inlier ratio increases strongly with confidence on both datasets, with the highest-confidence correspondences being correct in the large majority of cases, making confidence a reliable basis for weighting plane matches.

\subsection{Global Registration Performance}
For the main registration benchmarks, methods that operate on a supplied correspondence set are given the same point correspondences, obtained using FPFH descriptors followed by mutual nearest-neighbor (MNN) matching.
This includes RANSAC, FGR, TEASER++, MAC, CLIPPER+, PointDSC, and VBReg.
The only exception is MAC: its input correspondence set is capped at 4,000 matches when necessary, as larger sets exceed 32\,GB of memory.
For methods requiring a noise-bound parameter, we use the voxel size $v$.
For QUATRO, following QUATRO++~\cite{lim2024quatropp}, we segment the ground on KITTI and ETH using TRAVEL~\cite{oh2022travel}, remove the identified ground points, and apply the same point-matching frontend to the remaining points.
PARTE similarly uses TRAVEL for ground segmentation in KITTI, but it keeps the segmented ground for plane-matching.
Methods that include their own feature extraction or correspondence-generation stage are evaluated using their native pipelines, including PREDATOR, G3Reg, KISS-Matcher, VPFBR, and PLADE.
Unless otherwise stated, reported runtimes include the complete computational pipeline from feature extraction and matching through outlier rejection and transformation estimation, excluding file I/O.

\textbf{Indoor RGB-D registration.}
We first evaluate global registration on 3DMatch and the more challenging 3DLoMatch benchmark.
Table~\ref{tab:indoor-registration} reports the results using FPFH computed with camera-oriented normals.
PARTE achieves success rates of $89.6\%$ on 3DMatch and $55.8\%$ on 3DLoMatch, ranking second only to PREDATOR on both benchmarks.
PARTE also obtains the lowest rotation error on both datasets and the lowest translation error on 3DMatch, while maintaining relatively low end-to-end runtime.
The improvement is particularly notable on 3DLoMatch, where the low overlap produces a substantially more difficult point-matching problem.

\begin{table}[th]
\centering
\scriptsize
\caption{Indoor registration performance on 3DMatch and 3DLoMatch using
FPFH descriptors with mutual nearest-neighbor matching.
Reported runtimes include the complete registration pipeline, from feature computation and matching through outlier rejection and transformation estimation.}
\label{tab:indoor-registration}
\setlength{\tabcolsep}{2pt}
\renewcommand{\arraystretch}{0.9}
\resizebox{\columnwidth}{!}{%
\begin{tabular}{lcccc@{\hspace{6pt}}cccc}
\toprule
\multirow{2}{*}{Method} & \multicolumn{4}{c}{3DMatch} & \multicolumn{4}{c}{3DLoMatch} \\
\cmidrule(lr){2-5}\cmidrule(lr){6-9}
 & SR & ATE & ARE & Time & SR & ATE & ARE & Time \\
\midrule
\multicolumn{9}{l}{\textit{Baseline}} \\[-1pt]
RANSAC-$10^{3}$ & \cellcolor{ResultYellow!80!ResultGreen}59.8 & \cellcolor{ResultYellow!63!ResultGreen}9.5 & \cellcolor{ResultYellow!45!ResultGreen}3.4 & \cellcolor{ResultYellow!45!ResultGreen}63.4 & \cellcolor{ResultRed!68!ResultYellow}15.8 & \cellcolor{ResultYellow!90!ResultGreen}13.5 & \cellcolor{ResultYellow!66!ResultGreen}5.0 & \cellcolor{ResultYellow!50!ResultGreen}58.7 \\
RANSAC-$10^{5}$ & \cellcolor{ResultYellow!29!ResultGreen}85.6 & \cellcolor{ResultYellow!56!ResultGreen}8.4 & \cellcolor{ResultYellow!37!ResultGreen}2.8 & \cellcolor{ResultYellow!54!ResultGreen}70.8 & \cellcolor{ResultRed!8!ResultYellow}45.8 & \cellcolor{ResultYellow!88!ResultGreen}13.2 & \cellcolor{ResultYellow!60!ResultGreen}4.5 & \cellcolor{ResultYellow!66!ResultGreen}70.3 \\
RANSAC-$10^{7}$ & \cellcolor{ResultYellow!25!ResultGreen}87.7 & \cellcolor{ResultYellow!56!ResultGreen}8.4 & \cellcolor{ResultYellow!37!ResultGreen}2.8 & \cellcolor{ResultRed!16!ResultYellow}120.2 & \cellcolor{ResultYellow!98!ResultGreen}50.9 & \cellcolor{ResultYellow!87!ResultGreen}13.0 & \cellcolor{ResultYellow!58!ResultGreen}4.4 & \cellcolor{ResultRed!100!ResultYellow}282.7 \\
FGR & \cellcolor{ResultYellow!39!ResultGreen}80.3 & \cellcolor{ResultYellow!45!ResultGreen}6.8 & \cellcolor{ResultYellow!30!ResultGreen}2.2 & \cellcolor{ResultYellow!81!ResultGreen}92.3 & \cellcolor{ResultRed!55!ResultYellow}22.6 & \cellcolor{ResultYellow!72!ResultGreen}\textbf{10.7} & \cellcolor{ResultYellow!49!ResultGreen}3.7 & \cellcolor{ResultYellow!62!ResultGreen}67.2 \\
\addlinespace[2pt]
\multicolumn{9}{l}{\textit{Robust point-based methods}} \\[-1pt]
KISS-Matcher & \cellcolor{ResultYellow!49!ResultGreen}75.5 & \cellcolor{ResultYellow!58!ResultGreen}8.7 & \cellcolor{ResultYellow!39!ResultGreen}2.9 & \cellcolor{ResultRed!4!ResultYellow}110.8 & \cellcolor{ResultRed!46!ResultYellow}26.8 & \cellcolor{ResultYellow!87!ResultGreen}13.0 & \cellcolor{ResultYellow!61!ResultGreen}4.5 & \cellcolor{ResultYellow!3!ResultGreen}\textbf{26.4} \\
CLIPPER+ & \cellcolor{ResultYellow!28!ResultGreen}85.9 & \cellcolor{ResultYellow!46!ResultGreen}7.0 & \cellcolor{ResultYellow!30!ResultGreen}2.3 & \cellcolor{ResultRed!12!ResultYellow}117.5 & \cellcolor{ResultRed!11!ResultYellow}44.7 & \cellcolor{ResultYellow!76!ResultGreen}11.4 & \cellcolor{ResultYellow!52!ResultGreen}3.9 & \cellcolor{ResultRed!52!ResultYellow}130.1 \\
TEASER++ & \cellcolor{ResultYellow!27!ResultGreen}86.6 & \cellcolor{ResultYellow!48!ResultGreen}7.1 & \cellcolor{ResultYellow!30!ResultGreen}2.3 & \cellcolor{ResultYellow!63!ResultGreen}77.7 & \cellcolor{ResultRed!4!ResultYellow}48.0 & \cellcolor{ResultYellow!77!ResultGreen}11.6 & \cellcolor{ResultYellow!51!ResultGreen}3.9 & \cellcolor{ResultYellow!58!ResultGreen}64.5 \\
MAC & \cellcolor{ResultYellow!24!ResultGreen}88.0 & \cellcolor{ResultYellow!45!ResultGreen}6.8 & \cellcolor{ResultYellow!28!ResultGreen}2.1 & \cellcolor{ResultRed!6!ResultYellow}112.4 & \cellcolor{ResultYellow!95!ResultGreen}52.4 & \cellcolor{ResultYellow!77!ResultGreen}11.6 & \cellcolor{ResultYellow!49!ResultGreen}3.7 & \cellcolor{ResultYellow!88!ResultGreen}85.3 \\
\addlinespace[2pt]
\multicolumn{9}{l}{\textit{Learning-based methods}} \\[-1pt]
PointDSC & \cellcolor{ResultYellow!29!ResultGreen}85.5 & \cellcolor{ResultYellow!44!ResultGreen}\underline{6.6} & \cellcolor{ResultYellow!29!ResultGreen}2.1 & \cellcolor{ResultYellow!56!ResultGreen}72.2 & \cellcolor{ResultRed!17!ResultYellow}41.5 & \cellcolor{ResultYellow!73!ResultGreen}11.0 & \cellcolor{ResultYellow!48!ResultGreen}3.6 & \cellcolor{ResultYellow!59!ResultGreen}65.4 \\
VBReg & \cellcolor{ResultYellow!21!ResultGreen}89.4 & \cellcolor{ResultYellow!45!ResultGreen}6.8 & \cellcolor{ResultYellow!29!ResultGreen}2.1 & \cellcolor{ResultYellow!94!ResultGreen}102.8 & \cellcolor{ResultYellow!93!ResultGreen}53.5 & \cellcolor{ResultYellow!77!ResultGreen}11.6 & \cellcolor{ResultYellow!50!ResultGreen}3.7 & \cellcolor{ResultRed!0!ResultYellow}94.0 \\
PREDATOR & \cellcolor{ResultYellow!16!ResultGreen}\textbf{92.1} & \cellcolor{ResultYellow!49!ResultGreen}7.4 & \cellcolor{ResultYellow!30!ResultGreen}2.3 & \cellcolor{ResultYellow!85!ResultGreen}95.7 & \cellcolor{ResultYellow!78!ResultGreen}\textbf{61.0} & \cellcolor{ResultYellow!73!ResultGreen}11.0 & \cellcolor{ResultYellow!49!ResultGreen}3.7 & \cellcolor{ResultYellow!83!ResultGreen}82.0 \\
\addlinespace[2pt]
\multicolumn{9}{l}{\textit{Segment- and plane-based methods}} \\[-1pt]
VPFBR & \cellcolor{ResultRed!47!ResultYellow}26.5 & \cellcolor{ResultYellow!87!ResultGreen}13.0 & \cellcolor{ResultYellow!38!ResultGreen}2.9 & \cellcolor{ResultYellow!2!ResultGreen}\textbf{28.8} & \cellcolor{ResultRed!94!ResultYellow}2.8 & \cellcolor{ResultRed!8!ResultYellow}16.3 & \cellcolor{ResultYellow!49!ResultGreen}3.7 & \cellcolor{ResultYellow!3!ResultGreen}\underline{26.4} \\
PLADE & \cellcolor{ResultRed!31!ResultYellow}34.4 & \cellcolor{ResultYellow!55!ResultGreen}8.2 & \cellcolor{ResultYellow!27!ResultGreen}\underline{2.0} & \cellcolor{ResultRed!100!ResultYellow}1664.7 & \cellcolor{ResultRed!85!ResultYellow}7.3 & \cellcolor{ResultYellow!94!ResultGreen}14.2 & \cellcolor{ResultYellow!42!ResultGreen}\textbf{3.2} & \cellcolor{ResultRed!100!ResultYellow}1292.1 \\
G3Reg & \cellcolor{ResultRed!90!ResultYellow}4.8 & \cellcolor{ResultYellow!95!ResultGreen}14.2 & \cellcolor{ResultYellow!83!ResultGreen}6.2 & \cellcolor{ResultYellow!10!ResultGreen}\underline{35.1} & \cellcolor{ResultRed!100!ResultYellow}0.0 & \cellcolor{ResultRed!100!ResultYellow}-- & \cellcolor{ResultRed!100!ResultYellow}-- & \cellcolor{ResultYellow!27!ResultGreen}42.9 \\
\addlinespace[2pt]
\multicolumn{9}{l}{\textit{Ours}} \\[-1pt]
\textbf{PARTE} & \cellcolor{ResultYellow!21!ResultGreen}\underline{89.6} & \cellcolor{ResultYellow!42!ResultGreen}\textbf{6.2} & \cellcolor{ResultYellow!25!ResultGreen}\textbf{1.8} & \cellcolor{ResultYellow!40!ResultGreen}59.4 & \cellcolor{ResultYellow!88!ResultGreen}\underline{55.8} & \cellcolor{ResultYellow!72!ResultGreen}\underline{10.8} & \cellcolor{ResultYellow!45!ResultGreen}\underline{3.4} & \cellcolor{ResultYellow!47!ResultGreen}56.8 \\
\bottomrule
\end{tabular}%

}
\end{table}

The quality of the point correspondences has a substantial effect on registration performance.
In particular, consistently orienting the point normals toward the camera considerably improves FPFH matching compared with independently estimated unoriented normals.
Table~\ref{tab:indoor-frontends} compares several point-matching frontends to separate this effect from the registration method itself.
PARTE achieves the highest success rate with both FPFH variants on 3DMatch and 3DLoMatch, and remains the highest-performing non-learned method when FCGF correspondences are matched using MNN.
When 1,000 FCGF correspondences are retained, the gap narrows and MAC performs best among the geometric methods.
This correspondence set contains more competing mutually consistent structures, a setting that MAC explicitly addresses by considering multiple maximal cliques rather than relying on a single maximum clique.
Overall, PARTE remains among the top-performing methods across the different
point frontends, indicating that its performance is not tied to a particular
point descriptor or matching strategy.

\begin{table}[t]
\centering
\caption{Success-rate comparison under different point-matching frontends on
3DMatch and 3DLoMatch. Oriented FPFH uses camera-oriented normals, while FPFH
uses independently estimated unoriented normals. All frontends use mutual
nearest-neighbor matching, except FCGF + 1K, which uses 1,000 sampled
correspondences.}
\label{tab:indoor-frontends}
\scriptsize
\setlength{\tabcolsep}{2pt}
\renewcommand{\arraystretch}{0.9}
\resizebox{\columnwidth}{!}{%
\begin{tabular}{lcc!{\color{white}\vrule width 5pt}cc!{\color{white}\vrule width 5pt}cc!{\color{white}\vrule width 5pt}cc}
\toprule
\multirow{2}{*}{Method} & \multicolumn{2}{c}{Oriented FPFH} & \multicolumn{2}{c}{FPFH} & \multicolumn{2}{c}{FCGF} & \multicolumn{2}{c}{FCGF + 1K} \\
\cmidrule(lr){2-3}\cmidrule(lr){4-5}\cmidrule(lr){6-7}\cmidrule(lr){8-9}
 & 3DM & 3DLo & 3DM & 3DLo & 3DM & 3DLo & 3DM & 3DLo \\
\midrule
\multicolumn{9}{l}{\textit{Baseline}} \\[-1pt]
RANSAC-$10^{3}$ & \cellcolor{ResultYellow!80!ResultGreen}59.8 & \cellcolor{ResultRed!68!ResultYellow}15.8 & \cellcolor{ResultRed!15!ResultYellow}42.3 & \cellcolor{ResultRed!91!ResultYellow}4.4 & \cellcolor{ResultYellow!27!ResultGreen}86.3 & \cellcolor{ResultRed!24!ResultYellow}37.8 & \cellcolor{ResultYellow!38!ResultGreen}80.8 & \cellcolor{ResultRed!61!ResultYellow}19.4 \\
RANSAC-$10^{5}$ & \cellcolor{ResultYellow!29!ResultGreen}85.6 & \cellcolor{ResultRed!8!ResultYellow}45.8 & \cellcolor{ResultYellow!53!ResultGreen}73.4 & \cellcolor{ResultRed!50!ResultYellow}25.2 & \cellcolor{ResultYellow!23!ResultGreen}88.6 & \cellcolor{ResultRed!11!ResultYellow}44.6 & \cellcolor{ResultYellow!20!ResultGreen}90.0 & \cellcolor{ResultRed!5!ResultYellow}47.5 \\
RANSAC-$10^{7}$ & \cellcolor{ResultYellow!25!ResultGreen}87.7 & \cellcolor{ResultYellow!98!ResultGreen}50.9 & \cellcolor{ResultYellow!47!ResultGreen}76.3 & \cellcolor{ResultRed!38!ResultYellow}31.2 & \cellcolor{ResultYellow!23!ResultGreen}88.6 & \cellcolor{ResultRed!11!ResultYellow}44.5 & \cellcolor{ResultYellow!19!ResultGreen}90.4 & \cellcolor{ResultYellow!98!ResultGreen}50.9 \\
FGR & \cellcolor{ResultYellow!39!ResultGreen}80.3 & \cellcolor{ResultRed!55!ResultYellow}22.6 & \cellcolor{ResultYellow!67!ResultGreen}66.6 & \cellcolor{ResultRed!79!ResultYellow}10.5 & \cellcolor{ResultYellow!23!ResultGreen}88.4 & \cellcolor{ResultRed!24!ResultYellow}37.8 & \cellcolor{ResultYellow!25!ResultGreen}87.3 & \cellcolor{ResultRed!39!ResultYellow}30.7 \\
\addlinespace[2pt]
\multicolumn{9}{l}{\textit{Graph-based robust estimators}} \\[-1pt]
CLIPPER+ & \cellcolor{ResultYellow!28!ResultGreen}85.9 & \cellcolor{ResultRed!11!ResultYellow}44.7 & \cellcolor{ResultYellow!49!ResultGreen}75.5 & \cellcolor{ResultRed!50!ResultYellow}24.9 & \cellcolor{ResultYellow!25!ResultGreen}87.4 & \cellcolor{ResultRed!16!ResultYellow}41.9 & \cellcolor{ResultYellow!23!ResultGreen}88.7 & \cellcolor{ResultRed!10!ResultYellow}45.1 \\
TEASER++ & \cellcolor{ResultYellow!27!ResultGreen}86.6 & \cellcolor{ResultRed!4!ResultYellow}48.0 & \cellcolor{ResultYellow!47!ResultGreen}76.3 & \cellcolor{ResultRed!44!ResultYellow}27.8 & \cellcolor{ResultYellow!29!ResultGreen}85.6 & \cellcolor{ResultRed!23!ResultYellow}38.5 & \cellcolor{ResultYellow!26!ResultGreen}87.2 & \cellcolor{ResultRed!14!ResultYellow}43.2 \\
MAC & \cellcolor{ResultYellow!24!ResultGreen}88.0 & \cellcolor{ResultYellow!95!ResultGreen}52.4 & \cellcolor{ResultYellow!45!ResultGreen}77.4 & \cellcolor{ResultRed!36!ResultYellow}32.1 & \cellcolor{ResultYellow!29!ResultGreen}85.3 & \cellcolor{ResultRed!10!ResultYellow}44.9 & \cellcolor{ResultYellow!17!ResultGreen}\underline{91.7} & \cellcolor{ResultYellow!91!ResultGreen}\underline{54.6} \\
\addlinespace[2pt]
\multicolumn{9}{l}{\textit{Learning-based methods}} \\[-1pt]
PointDSC & \cellcolor{ResultYellow!29!ResultGreen}85.5 & \cellcolor{ResultRed!17!ResultYellow}41.5 & \cellcolor{ResultYellow!56!ResultGreen}71.8 & \cellcolor{ResultRed!56!ResultYellow}22.0 & \cellcolor{ResultYellow!23!ResultGreen}88.7 & \cellcolor{ResultRed!12!ResultYellow}43.9 & \cellcolor{ResultYellow!18!ResultGreen}91.1 & \cellcolor{ResultRed!6!ResultYellow}47.1 \\
VBReg & \cellcolor{ResultYellow!21!ResultGreen}\underline{89.4} & \cellcolor{ResultYellow!93!ResultGreen}\underline{53.5} & \cellcolor{ResultYellow!41!ResultGreen}\underline{79.7} & \cellcolor{ResultRed!33!ResultYellow}\underline{33.6} & \cellcolor{ResultYellow!19!ResultGreen}\textbf{90.6} & \cellcolor{ResultRed!3!ResultYellow}\textbf{48.7} & \cellcolor{ResultYellow!16!ResultGreen}\textbf{92.2} & \cellcolor{ResultYellow!90!ResultGreen}\textbf{55.2} \\
\addlinespace[2pt]
\multicolumn{9}{l}{\textit{Ours}} \\[-1pt]
\textbf{PARTE} & \cellcolor{ResultYellow!21!ResultGreen}\textbf{89.6} & \cellcolor{ResultYellow!88!ResultGreen}\textbf{55.8} & \cellcolor{ResultYellow!29!ResultGreen}\textbf{85.5} & \cellcolor{ResultRed!22!ResultYellow}\textbf{39.1} & \cellcolor{ResultYellow!21!ResultGreen}\underline{89.3} & \cellcolor{ResultRed!5!ResultYellow}\underline{47.7} & \cellcolor{ResultYellow!20!ResultGreen}90.2 & \cellcolor{ResultYellow!99!ResultGreen}50.5 \\
\bottomrule
\end{tabular}%

}
\end{table}

The frontend comparison also illustrates the trade-off between correspondence quantity and robust-estimation cost.
MNN produces a relatively compact correspondence set, whereas retaining a larger number of FCGF matches substantially increases the runtime of several robust estimators.
This motivates the use of MNN in the main evaluation, while the consistent performance of PARTE across the different frontends indicates that its registration gains are largely complementary to the choice of point-matching method.

\textbf{Outdoor LiDAR registration.}
We evaluate outdoor registration on the standard KITTI-10m benchmark and the more challenging KITTI-LC loop-closure benchmark.

\begin{table}[th]
\centering
\caption{Registration performance on the KITTI-10m benchmark.
Success rate is reported for sequences 08--10 and overall, together with
mean end-to-end runtime.}
\label{tab:kitti10m}
\scriptsize
\setlength{\tabcolsep}{2pt}
\resizebox{\columnwidth}{!}{%
\begin{tabular}{lrrrr!{\color{white}\vrule width 2.8pt}rrrr}
\toprule
& \multicolumn{4}{c}{Success Rate (\%)} & \multicolumn{4}{c}{Runtime (ms)} \\
\cmidrule(lr){2-5} \cmidrule(lr){6-9}
Method & 08 & 09 & 10 & All & 08 & 09 & 10 & All \\
\midrule
\multicolumn{9}{l}{\textit{Baseline}} \\[-1pt]
RANSAC-$10^{3}$ & \cellcolor{ResultRed!90!ResultYellow}4.9 & \cellcolor{ResultRed!67!ResultYellow}16.7 & \cellcolor{ResultRed!66!ResultYellow}17.2 & \cellcolor{ResultRed!79!ResultYellow}10.3 & \cellcolor{ResultYellow!82!ResultGreen}419.2 & \cellcolor{ResultYellow!74!ResultGreen}432.1 & \cellcolor{ResultYellow!67!ResultGreen}287.0 & \cellcolor{ResultYellow!77!ResultGreen}402.3 \\
RANSAC-$10^{5}$ & \cellcolor{ResultYellow!21!ResultGreen}89.3 & \cellcolor{ResultYellow!11!ResultGreen}94.4 & \cellcolor{ResultYellow!11!ResultGreen}94.3 & \cellcolor{ResultYellow!17!ResultGreen}91.5 & \cellcolor{ResultYellow!86!ResultGreen}434.2 & \cellcolor{ResultYellow!77!ResultGreen}446.4 & \cellcolor{ResultYellow!72!ResultGreen}300.0 & \cellcolor{ResultYellow!80!ResultGreen}416.7 \\
RANSAC-$10^{7}$ & \cellcolor{ResultYellow!5!ResultGreen}97.4 & \cellcolor{ResultYellow!0!ResultGreen}\textbf{100.0} & \cellcolor{ResultYellow!5!ResultGreen}97.7 & \cellcolor{ResultYellow!4!ResultGreen}98.2 & \cellcolor{ResultRed!28!ResultYellow}594.3 & \cellcolor{ResultRed!3!ResultYellow}562.0 & \cellcolor{ResultYellow!94!ResultGreen}367.1 & \cellcolor{ResultRed!14!ResultYellow}549.3 \\
FGR & \cellcolor{ResultRed!55!ResultYellow}22.5 & \cellcolor{ResultRed!32!ResultYellow}34.0 & \cellcolor{ResultRed!24!ResultYellow}37.9 & \cellcolor{ResultRed!44!ResultYellow}28.2 & \cellcolor{ResultYellow!98!ResultGreen}480.1 & \cellcolor{ResultYellow!91!ResultGreen}506.6 & \cellcolor{ResultYellow!87!ResultGreen}345.7 & \cellcolor{ResultYellow!93!ResultGreen}466.8 \\
\addlinespace[1.5pt]
\multicolumn{9}{l}{\textit{Robust point-based methods}} \\[-1pt]
KISS-Matcher & \cellcolor{ResultYellow!6!ResultGreen}97.1 & \cellcolor{ResultYellow!2!ResultGreen}98.8 & \cellcolor{ResultYellow!11!ResultGreen}94.3 & \cellcolor{ResultYellow!6!ResultGreen}97.1 & \cellcolor{ResultYellow!0!ResultGreen}\underline{98.8} & \cellcolor{ResultYellow!2!ResultGreen}\underline{107.1} & \cellcolor{ResultYellow!0!ResultGreen}\textbf{78.4} & \cellcolor{ResultYellow!0!ResultGreen}\underline{98.0} \\
CLIPPER+ & \cellcolor{ResultYellow!4!ResultGreen}98.0 & \cellcolor{ResultYellow!1!ResultGreen}\underline{99.4} & \cellcolor{ResultYellow!0!ResultGreen}\textbf{100.0} & \cellcolor{ResultYellow!3!ResultGreen}98.7 & \cellcolor{ResultRed!81!ResultYellow}800.1 & \cellcolor{ResultRed!100!ResultYellow}1487.3 & \cellcolor{ResultRed!85!ResultYellow}632.3 & \cellcolor{ResultRed!100!ResultYellow}974.1 \\
TEASER++ & \cellcolor{ResultYellow!4!ResultGreen}98.0 & \cellcolor{ResultYellow!0!ResultGreen}\textbf{100.0} & \cellcolor{ResultYellow!0!ResultGreen}\textbf{100.0} & \cellcolor{ResultYellow!2!ResultGreen}98.9 & \cellcolor{ResultRed!12!ResultYellow}534.3 & \cellcolor{ResultRed!5!ResultYellow}574.1 & \cellcolor{ResultYellow!95!ResultGreen}367.4 & \cellcolor{ResultRed!7!ResultYellow}519.8 \\
QUATRO & \cellcolor{ResultYellow!2!ResultGreen}\underline{99.0} & \cellcolor{ResultYellow!4!ResultGreen}98.1 & \cellcolor{ResultYellow!9!ResultGreen}95.4 & \cellcolor{ResultYellow!4!ResultGreen}98.2 & \cellcolor{ResultYellow!63!ResultGreen}345.2 & \cellcolor{ResultYellow!58!ResultGreen}359.6 & \cellcolor{ResultYellow!52!ResultGreen}242.1 & \cellcolor{ResultYellow!59!ResultGreen}333.3 \\
MAC & \cellcolor{ResultYellow!6!ResultGreen}97.1 & \cellcolor{ResultYellow!0!ResultGreen}\textbf{100.0} & \cellcolor{ResultYellow!0!ResultGreen}\textbf{100.0} & \cellcolor{ResultYellow!3!ResultGreen}98.4 & \cellcolor{ResultRed!100!ResultYellow}1967.2 & \cellcolor{ResultRed!100!ResultYellow}2053.1 & \cellcolor{ResultRed!100!ResultYellow}1433.8 & \cellcolor{ResultRed!100!ResultYellow}1908.7 \\
\addlinespace[1.5pt]
\multicolumn{9}{l}{\textit{Learning-based methods}} \\[-1pt]
PointDSC & \cellcolor{ResultYellow!8!ResultGreen}96.1 & \cellcolor{ResultYellow!0!ResultGreen}\textbf{100.0} & \cellcolor{ResultYellow!2!ResultGreen}\underline{98.9} & \cellcolor{ResultYellow!5!ResultGreen}97.7 & \cellcolor{ResultYellow!88!ResultGreen}441.6 & \cellcolor{ResultYellow!80!ResultGreen}457.0 & \cellcolor{ResultYellow!73!ResultGreen}303.5 & \cellcolor{ResultYellow!82!ResultGreen}424.5 \\
VBReg & \cellcolor{ResultYellow!7!ResultGreen}96.7 & \cellcolor{ResultYellow!2!ResultGreen}98.8 & \cellcolor{ResultYellow!5!ResultGreen}97.7 & \cellcolor{ResultYellow!5!ResultGreen}97.5 & \cellcolor{ResultRed!7!ResultYellow}514.1 & \cellcolor{ResultYellow!98!ResultGreen}538.2 & \cellcolor{ResultYellow!91!ResultGreen}356.1 & \cellcolor{ResultRed!1!ResultYellow}496.4 \\
PREDATOR & \cellcolor{ResultYellow!1!ResultGreen}\textbf{99.7} & \cellcolor{ResultYellow!0!ResultGreen}\textbf{100.0} & \cellcolor{ResultYellow!0!ResultGreen}\textbf{100.0} & \cellcolor{ResultYellow!0!ResultGreen}\textbf{99.8} & \cellcolor{ResultRed!20!ResultYellow}564.5 & \cellcolor{ResultRed!7!ResultYellow}579.2 & \cellcolor{ResultRed!28!ResultYellow}465.7 & \cellcolor{ResultRed!15!ResultYellow}553.3 \\
\addlinespace[1.5pt]
\multicolumn{9}{l}{\textit{Segment- and plane-based methods}} \\[-1pt]
VPFBR & \cellcolor{ResultRed!31!ResultYellow}34.5 & \cellcolor{ResultRed!40!ResultYellow}30.2 & \cellcolor{ResultRed!52!ResultYellow}24.1 & \cellcolor{ResultRed!37!ResultYellow}31.7 & \cellcolor{ResultYellow!22!ResultGreen}186.7 & \cellcolor{ResultYellow!10!ResultGreen}143.1 & \cellcolor{ResultYellow!12!ResultGreen}125.6 & \cellcolor{ResultYellow!17!ResultGreen}165.5 \\
PLADE & \cellcolor{ResultRed!96!ResultYellow}2.0 & \cellcolor{ResultRed!100!ResultYellow}0.0 & \cellcolor{ResultRed!100!ResultYellow}0.0 & \cellcolor{ResultRed!98!ResultYellow}1.1 & \cellcolor{ResultRed!81!ResultYellow}799.9 & \cellcolor{ResultRed!40!ResultYellow}730.6 & \cellcolor{ResultRed!100!ResultYellow}691.1 & \cellcolor{ResultRed!68!ResultYellow}762.7 \\
G3Reg & \cellcolor{ResultYellow!2!ResultGreen}\underline{99.0} & \cellcolor{ResultYellow!2!ResultGreen}98.8 & \cellcolor{ResultYellow!0!ResultGreen}\textbf{100.0} & \cellcolor{ResultYellow!2!ResultGreen}\underline{99.1} & \cellcolor{ResultYellow!0!ResultGreen}\textbf{90.1} & \cellcolor{ResultYellow!0!ResultGreen}\textbf{85.1} & \cellcolor{ResultYellow!0!ResultGreen}\underline{85.9} & \cellcolor{ResultYellow!0!ResultGreen}\textbf{88.0} \\
\addlinespace[1.5pt]
\multicolumn{9}{l}{\textit{Ours}} \\[-1pt]
\textbf{PARTE} & \cellcolor{ResultYellow!1!ResultGreen}\textbf{99.7} & \cellcolor{ResultYellow!0!ResultGreen}\textbf{100.0} & \cellcolor{ResultYellow!0!ResultGreen}\textbf{100.0} & \cellcolor{ResultYellow!0!ResultGreen}\textbf{99.8} & \cellcolor{ResultYellow!21!ResultGreen}182.9 & \cellcolor{ResultYellow!20!ResultGreen}188.3 & \cellcolor{ResultYellow!18!ResultGreen}141.5 & \cellcolor{ResultYellow!20!ResultGreen}178.0 \\
\bottomrule
\end{tabular}%

}
\end{table}

Table~\ref{tab:kitti10m} reports the results on KITTI-10m. 
PARTE achieves a $99.8\%$ overall success rate, matching PREDATOR for the highest performance and reaching near-perfect registration on all three sequences. 
G3Reg also performs strongly with a $99.1\%$ success rate, while most modern robust registration methods exceed $97\%$. These results indicate that KITTI-10m is largely saturated by current global registration methods~\cite{qiao2024g3reg}. 
PARTE nevertheless reaches this performance with a mean runtime of $178$ ms, substantially below PREDATOR, TEASER++, CLIPPER+, and MAC, although G3Reg and KISS-Matcher remain faster. 
The RANSAC results also show a clear convergence trend, increasing from $10.3\%$ success at $10^3$ iterations to $91.5\%$ at $10^5$ and $98.2\%$ at $10^7$ iterations.

\begin{table*}[t]
\centering
\caption{KITTI-LC registration performance across sequences and translation ranges. We report success rate under the $2$ m/$5^\circ$ criterion and mean end-to-end runtime over all evaluated pairs.}
\label{tab:kittilc-registration}
\scriptsize
\setlength{\tabcolsep}{1.6pt}
\setlength{\fboxsep}{0.7pt}
\renewcommand{\arraystretch}{0.94}
\resizebox{\textwidth}{!}{%
\begin{tabular}{lrrrrr!{\color{white}\vrule width 3pt}rrrrr!{\color{white}\vrule width 3pt}rrrrr!{\color{white}\vrule width 3pt}rrrrr!{\color{white}\vrule width 3pt}r}
\toprule
\multirow{2}{*}{Method} & \multicolumn{5}{c}{0--10 m} & \multicolumn{5}{c}{10--20 m} & \multicolumn{5}{c}{20--30 m} & \multicolumn{5}{c}{All} & \multicolumn{1}{c}{Runtime} \\
\cmidrule(lr){2-6}\cmidrule(lr){7-11}\cmidrule(lr){12-16}\cmidrule(lr){17-21}
 & 00 & 02 & 05 & 06 & 08 & 00 & 02 & 05 & 06 & 08 & 00 & 02 & 05 & 06 & 08 & 00 & 02 & 05 & 06 & 08 & (ms) \\
\midrule
\multicolumn{22}{l}{\textit{Baseline}} \\[-1pt]
RANSAC-$10^{3}$ & \cellcolor{ResultRed!8!ResultYellow}45.9 & \cellcolor{ResultRed!10!ResultYellow}44.8 & \cellcolor{ResultRed!36!ResultYellow}31.8 & \cellcolor{ResultRed!22!ResultYellow}38.8 & \cellcolor{ResultRed!66!ResultYellow}17.2 & \cellcolor{ResultRed!92!ResultYellow}3.8 & \cellcolor{ResultRed!93!ResultYellow}3.5 & \cellcolor{ResultRed!96!ResultYellow}2.0 & \cellcolor{ResultRed!97!ResultYellow}1.6 & \cellcolor{ResultRed!99!ResultYellow}0.6 & \cellcolor{ResultRed!100!ResultYellow}0.0 & \cellcolor{ResultRed!100!ResultYellow}0.0 & \cellcolor{ResultRed!100!ResultYellow}0.0 & \cellcolor{ResultRed!100!ResultYellow}0.0 & \cellcolor{ResultRed!100!ResultYellow}0.0 & \cellcolor{ResultRed!70!ResultYellow}14.8 & \cellcolor{ResultRed!72!ResultYellow}14.2 & \cellcolor{ResultRed!80!ResultYellow}10.1 & \cellcolor{ResultRed!83!ResultYellow}8.5 & \cellcolor{ResultRed!90!ResultYellow}5.0 & \cellcolor{ResultYellow!50!ResultGreen}360.1 \\
RANSAC-$10^{5}$ & \cellcolor{ResultYellow!1!ResultGreen}\underline{99.3} & \cellcolor{ResultYellow!19!ResultGreen}90.7 & \cellcolor{ResultYellow!1!ResultGreen}\underline{99.4} & \cellcolor{ResultYellow!0!ResultGreen}\textbf{100.0} & \cellcolor{ResultYellow!12!ResultGreen}\underline{94.0} & \cellcolor{ResultRed!5!ResultYellow}47.5 & \cellcolor{ResultRed!17!ResultYellow}41.4 & \cellcolor{ResultRed!23!ResultYellow}38.7 & \cellcolor{ResultRed!7!ResultYellow}46.6 & \cellcolor{ResultRed!35!ResultYellow}32.7 & \cellcolor{ResultRed!94!ResultYellow}2.9 & \cellcolor{ResultRed!86!ResultYellow}7.2 & \cellcolor{ResultRed!95!ResultYellow}2.3 & \cellcolor{ResultRed!80!ResultYellow}9.9 & \cellcolor{ResultRed!96!ResultYellow}2.2 & \cellcolor{ResultRed!8!ResultYellow}46.2 & \cellcolor{ResultRed!14!ResultYellow}42.8 & \cellcolor{ResultRed!13!ResultYellow}43.3 & \cellcolor{ResultRed!14!ResultYellow}42.9 & \cellcolor{ResultRed!24!ResultYellow}38.0 & \cellcolor{ResultYellow!53!ResultGreen}376.9 \\
RANSAC-$10^{7}$ & \cellcolor{ResultYellow!0!ResultGreen}\textbf{100.0} & \cellcolor{ResultYellow!19!ResultGreen}90.7 & \cellcolor{ResultYellow!0!ResultGreen}\textbf{100.0} & \cellcolor{ResultYellow!0!ResultGreen}\textbf{100.0} & \cellcolor{ResultYellow!0!ResultGreen}\textbf{100.0} & \cellcolor{ResultYellow!58!ResultGreen}71.0 & \cellcolor{ResultYellow!80!ResultGreen}60.1 & \cellcolor{ResultYellow!57!ResultGreen}71.4 & \cellcolor{ResultYellow!10!ResultGreen}95.2 & \cellcolor{ResultYellow!40!ResultGreen}80.2 & \cellcolor{ResultRed!67!ResultYellow}16.5 & \cellcolor{ResultRed!46!ResultYellow}26.9 & \cellcolor{ResultRed!73!ResultYellow}13.7 & \cellcolor{ResultRed!5!ResultYellow}47.6 & \cellcolor{ResultRed!82!ResultYellow}9.2 & \cellcolor{ResultYellow!81!ResultGreen}59.3 & \cellcolor{ResultYellow!87!ResultGreen}56.5 & \cellcolor{ResultYellow!83!ResultGreen}58.6 & \cellcolor{ResultYellow!46!ResultGreen}77.2 & \cellcolor{ResultYellow!83!ResultGreen}58.4 & \cellcolor{ResultRed!84!ResultYellow}1073.8 \\
FGR & \cellcolor{ResultYellow!81!ResultGreen}59.4 & \cellcolor{ResultRed!22!ResultYellow}39.0 & \cellcolor{ResultRed!1!ResultYellow}49.4 & \cellcolor{ResultYellow!60!ResultGreen}69.8 & \cellcolor{ResultRed!93!ResultYellow}3.7 & \cellcolor{ResultRed!93!ResultYellow}3.5 & \cellcolor{ResultRed!97!ResultYellow}1.5 & \cellcolor{ResultRed!99!ResultYellow}0.5 & \cellcolor{ResultRed!98!ResultYellow}0.8 & \cellcolor{ResultRed!100!ResultYellow}0.0 & \cellcolor{ResultRed!100!ResultYellow}0.0 & \cellcolor{ResultRed!100!ResultYellow}0.0 & \cellcolor{ResultRed!100!ResultYellow}0.0 & \cellcolor{ResultRed!100!ResultYellow}0.0 & \cellcolor{ResultRed!100!ResultYellow}0.0 & \cellcolor{ResultRed!63!ResultYellow}18.7 & \cellcolor{ResultRed!76!ResultYellow}11.8 & \cellcolor{ResultRed!70!ResultYellow}14.8 & \cellcolor{ResultRed!71!ResultYellow}14.6 & \cellcolor{ResultRed!98!ResultYellow}1.0 & \cellcolor{ResultYellow!59!ResultGreen}410.9 \\
\addlinespace[2pt]
\multicolumn{22}{l}{\textit{Robust point-based methods}} \\[-1pt]
KISS-Matcher & \cellcolor{ResultYellow!0!ResultGreen}\textbf{100.0} & \cellcolor{ResultYellow!23!ResultGreen}88.4 & \cellcolor{ResultYellow!1!ResultGreen}\underline{99.4} & \cellcolor{ResultYellow!0!ResultGreen}\textbf{100.0} & \cellcolor{ResultYellow!0!ResultGreen}\textbf{100.0} & \cellcolor{ResultYellow!83!ResultGreen}58.7 & \cellcolor{ResultYellow!90!ResultGreen}55.1 & \cellcolor{ResultYellow!82!ResultGreen}58.8 & \cellcolor{ResultYellow!57!ResultGreen}71.7 & \cellcolor{ResultYellow!58!ResultGreen}71.0 & \cellcolor{ResultRed!82!ResultYellow}8.9 & \cellcolor{ResultRed!78!ResultYellow}10.8 & \cellcolor{ResultRed!87!ResultYellow}6.4 & \cellcolor{ResultRed!71!ResultYellow}14.7 & \cellcolor{ResultRed!89!ResultYellow}5.4 & \cellcolor{ResultYellow!95!ResultGreen}52.4 & \cellcolor{ResultRed!4!ResultYellow}48.1 & \cellcolor{ResultYellow!97!ResultGreen}51.5 & \cellcolor{ResultYellow!91!ResultGreen}54.7 & \cellcolor{ResultYellow!92!ResultGreen}53.8 & \cellcolor{ResultYellow!1!ResultGreen}\underline{101.1} \\
CLIPPER+ & \cellcolor{ResultYellow!0!ResultGreen}\textbf{100.0} & \cellcolor{ResultYellow!15!ResultGreen}92.4 & \cellcolor{ResultYellow!0!ResultGreen}\textbf{100.0} & \cellcolor{ResultYellow!0!ResultGreen}\textbf{100.0} & \cellcolor{ResultYellow!0!ResultGreen}\textbf{100.0} & \cellcolor{ResultYellow!44!ResultGreen}78.0 & \cellcolor{ResultYellow!86!ResultGreen}57.1 & \cellcolor{ResultYellow!40!ResultGreen}79.9 & \cellcolor{ResultYellow!4!ResultGreen}98.0 & \cellcolor{ResultYellow!22!ResultGreen}88.9 & \cellcolor{ResultRed!63!ResultYellow}18.6 & \cellcolor{ResultRed!56!ResultYellow}22.0 & \cellcolor{ResultRed!72!ResultYellow}14.2 & \cellcolor{ResultYellow!76!ResultGreen}61.9 & \cellcolor{ResultRed!64!ResultYellow}17.8 & \cellcolor{ResultYellow!75!ResultGreen}62.4 & \cellcolor{ResultYellow!92!ResultGreen}54.1 & \cellcolor{ResultYellow!77!ResultGreen}61.6 & \cellcolor{ResultYellow!32!ResultGreen}84.0 & \cellcolor{ResultYellow!71!ResultGreen}64.7 & \cellcolor{ResultRed!100!ResultYellow}2574.9 \\
TEASER++ & \cellcolor{ResultYellow!0!ResultGreen}\textbf{100.0} & \cellcolor{ResultYellow!14!ResultGreen}\underline{93.0} & \cellcolor{ResultYellow!0!ResultGreen}\textbf{100.0} & \cellcolor{ResultYellow!0!ResultGreen}\textbf{100.0} & \cellcolor{ResultYellow!0!ResultGreen}\textbf{100.0} & \cellcolor{ResultYellow!37!ResultGreen}81.5 & \cellcolor{ResultYellow!77!ResultGreen}61.6 & \cellcolor{ResultYellow!27!ResultGreen}86.4 & \cellcolor{ResultYellow!2!ResultGreen}\underline{99.2} & \cellcolor{ResultYellow!9!ResultGreen}95.7 & \cellcolor{ResultRed!49!ResultYellow}25.7 & \cellcolor{ResultRed!38!ResultYellow}30.9 & \cellcolor{ResultRed!58!ResultYellow}21.0 & \cellcolor{ResultYellow!49!ResultGreen}75.4 & \cellcolor{ResultRed!47!ResultYellow}26.5 & \cellcolor{ResultYellow!68!ResultGreen}66.2 & \cellcolor{ResultYellow!82!ResultGreen}59.2 & \cellcolor{ResultYellow!67!ResultGreen}66.3 & \cellcolor{ResultYellow!20!ResultGreen}89.9 & \cellcolor{ResultYellow!59!ResultGreen}70.3 & \cellcolor{ResultYellow!75!ResultGreen}494.7 \\
QUATRO & \cellcolor{ResultYellow!2!ResultGreen}99.0 & \cellcolor{ResultYellow!0!ResultGreen}\textbf{100.0} & \cellcolor{ResultYellow!0!ResultGreen}\textbf{100.0} & \cellcolor{ResultYellow!0!ResultGreen}\textbf{100.0} & \cellcolor{ResultYellow!0!ResultGreen}\textbf{100.0} & \cellcolor{ResultYellow!12!ResultGreen}\underline{93.8} & \cellcolor{ResultYellow!36!ResultGreen}\underline{81.8} & \cellcolor{ResultYellow!9!ResultGreen}95.5 & \cellcolor{ResultYellow!2!ResultGreen}98.8 & \cellcolor{ResultYellow!4!ResultGreen}\underline{98.1} & \cellcolor{ResultYellow!96!ResultGreen}52.2 & \cellcolor{ResultRed!7!ResultYellow}46.6 & \cellcolor{ResultYellow!96!ResultGreen}52.1 & \cellcolor{ResultYellow!12!ResultGreen}94.0 & \cellcolor{ResultYellow!78!ResultGreen}61.1 & \cellcolor{ResultYellow!40!ResultGreen}79.9 & \cellcolor{ResultYellow!52!ResultGreen}\underline{73.9} & \cellcolor{ResultYellow!38!ResultGreen}80.8 & \cellcolor{ResultYellow!6!ResultGreen}97.2 & \cellcolor{ResultYellow!31!ResultGreen}84.4 & \cellcolor{ResultYellow!39!ResultGreen}304.3 \\
MAC & \cellcolor{ResultYellow!0!ResultGreen}\textbf{100.0} & \cellcolor{ResultYellow!17!ResultGreen}91.3 & \cellcolor{ResultYellow!0!ResultGreen}\textbf{100.0} & \cellcolor{ResultYellow!0!ResultGreen}\textbf{100.0} & \cellcolor{ResultYellow!0!ResultGreen}\textbf{100.0} & \cellcolor{ResultYellow!46!ResultGreen}77.1 & \cellcolor{ResultYellow!80!ResultGreen}60.1 & \cellcolor{ResultYellow!42!ResultGreen}78.9 & \cellcolor{ResultYellow!6!ResultGreen}96.8 & \cellcolor{ResultYellow!23!ResultGreen}88.3 & \cellcolor{ResultRed!57!ResultYellow}21.5 & \cellcolor{ResultRed!36!ResultYellow}31.8 & \cellcolor{ResultRed!62!ResultYellow}19.2 & \cellcolor{ResultYellow!84!ResultGreen}57.9 & \cellcolor{ResultRed!65!ResultYellow}17.3 & \cellcolor{ResultYellow!74!ResultGreen}63.2 & \cellcolor{ResultYellow!83!ResultGreen}58.5 & \cellcolor{ResultYellow!74!ResultGreen}63.1 & \cellcolor{ResultYellow!36!ResultGreen}82.0 & \cellcolor{ResultYellow!72!ResultGreen}64.2 & \cellcolor{ResultRed!100!ResultYellow}1721.4 \\
\addlinespace[2pt]
\multicolumn{22}{l}{\textit{Learning-based methods}} \\[-1pt]
PointDSC & \cellcolor{ResultYellow!0!ResultGreen}\textbf{100.0} & \cellcolor{ResultYellow!21!ResultGreen}89.5 & \cellcolor{ResultYellow!0!ResultGreen}\textbf{100.0} & \cellcolor{ResultYellow!0!ResultGreen}\textbf{100.0} & \cellcolor{ResultYellow!0!ResultGreen}\textbf{100.0} & \cellcolor{ResultYellow!70!ResultGreen}65.1 & \cellcolor{ResultRed!5!ResultYellow}47.5 & \cellcolor{ResultYellow!67!ResultGreen}66.3 & \cellcolor{ResultYellow!14!ResultGreen}92.8 & \cellcolor{ResultYellow!64!ResultGreen}67.9 & \cellcolor{ResultRed!81!ResultYellow}9.4 & \cellcolor{ResultRed!76!ResultYellow}12.1 & \cellcolor{ResultRed!81!ResultYellow}9.6 & \cellcolor{ResultRed!25!ResultYellow}37.3 & \cellcolor{ResultRed!89!ResultYellow}5.4 & \cellcolor{ResultYellow!91!ResultGreen}54.7 & \cellcolor{ResultRed!7!ResultYellow}46.4 & \cellcolor{ResultYellow!89!ResultGreen}55.4 & \cellcolor{ResultYellow!56!ResultGreen}72.2 & \cellcolor{ResultYellow!94!ResultGreen}52.8 & \cellcolor{ResultYellow!60!ResultGreen}414.3 \\
VBReg & \cellcolor{ResultYellow!0!ResultGreen}\textbf{100.0} & \cellcolor{ResultYellow!24!ResultGreen}87.8 & \cellcolor{ResultYellow!0!ResultGreen}\textbf{100.0} & \cellcolor{ResultYellow!0!ResultGreen}\textbf{100.0} & \cellcolor{ResultYellow!0!ResultGreen}\textbf{100.0} & \cellcolor{ResultYellow!43!ResultGreen}78.3 & \cellcolor{ResultYellow!92!ResultGreen}54.0 & \cellcolor{ResultYellow!37!ResultGreen}81.4 & \cellcolor{ResultYellow!7!ResultGreen}96.4 & \cellcolor{ResultYellow!37!ResultGreen}81.5 & \cellcolor{ResultRed!56!ResultYellow}21.8 & \cellcolor{ResultRed!51!ResultYellow}24.7 & \cellcolor{ResultRed!69!ResultYellow}15.5 & \cellcolor{ResultRed!18!ResultYellow}40.9 & \cellcolor{ResultRed!74!ResultYellow}13.0 & \cellcolor{ResultYellow!73!ResultGreen}63.7 & \cellcolor{ResultYellow!94!ResultGreen}52.8 & \cellcolor{ResultYellow!75!ResultGreen}62.6 & \cellcolor{ResultYellow!50!ResultGreen}75.0 & \cellcolor{ResultYellow!79!ResultGreen}60.3 & \cellcolor{ResultYellow!69!ResultGreen}465.1 \\
PREDATOR & \cellcolor{ResultYellow!58!ResultGreen}71.0 & \cellcolor{ResultYellow!77!ResultGreen}61.6 & \cellcolor{ResultYellow!49!ResultGreen}75.6 & \cellcolor{ResultYellow!3!ResultGreen}\underline{98.4} & \cellcolor{ResultYellow!54!ResultGreen}73.1 & \cellcolor{ResultRed!37!ResultYellow}31.7 & \cellcolor{ResultRed!58!ResultYellow}21.2 & \cellcolor{ResultRed!44!ResultYellow}28.1 & \cellcolor{ResultRed!64!ResultYellow}17.9 & \cellcolor{ResultRed!72!ResultYellow}14.2 & \cellcolor{ResultRed!94!ResultYellow}3.1 & \cellcolor{ResultRed!96!ResultYellow}1.8 & \cellcolor{ResultRed!96!ResultYellow}1.8 & \cellcolor{ResultRed!97!ResultYellow}1.6 & \cellcolor{ResultRed!100!ResultYellow}0.0 & \cellcolor{ResultRed!35!ResultYellow}32.7 & \cellcolor{ResultRed!49!ResultYellow}25.6 & \cellcolor{ResultRed!35!ResultYellow}32.5 & \cellcolor{ResultRed!44!ResultYellow}27.8 & \cellcolor{ResultRed!50!ResultYellow}25.2 & \cellcolor{ResultYellow!51!ResultGreen}366.3 \\
\addlinespace[2pt]
\multicolumn{22}{l}{\textit{Segment- and plane-based methods}} \\[-1pt]
VPFBR & \cellcolor{ResultYellow!20!ResultGreen}90.1 & \cellcolor{ResultRed!26!ResultYellow}37.2 & \cellcolor{ResultYellow!36!ResultGreen}81.8 & \cellcolor{ResultRed!4!ResultYellow}48.1 & \cellcolor{ResultRed!7!ResultYellow}46.3 & \cellcolor{ResultYellow!69!ResultGreen}65.7 & \cellcolor{ResultRed!80!ResultYellow}10.1 & \cellcolor{ResultYellow!95!ResultGreen}52.3 & \cellcolor{ResultRed!51!ResultYellow}24.3 & \cellcolor{ResultRed!38!ResultYellow}30.9 & \cellcolor{ResultRed!51!ResultYellow}24.7 & \cellcolor{ResultRed!90!ResultYellow}4.9 & \cellcolor{ResultRed!53!ResultYellow}23.3 & \cellcolor{ResultRed!84!ResultYellow}7.9 & \cellcolor{ResultRed!63!ResultYellow}18.4 & \cellcolor{ResultYellow!85!ResultGreen}57.7 & \cellcolor{ResultRed!68!ResultYellow}16.0 & \cellcolor{ResultYellow!99!ResultGreen}50.3 & \cellcolor{ResultRed!55!ResultYellow}22.6 & \cellcolor{ResultRed!39!ResultYellow}30.4 & \cellcolor{ResultYellow!11!ResultGreen}154.7 \\
PLADE & \cellcolor{ResultRed!64!ResultYellow}17.8 & \cellcolor{ResultRed!99!ResultYellow}0.6 & \cellcolor{ResultRed!91!ResultYellow}4.5 & \cellcolor{ResultRed!84!ResultYellow}7.8 & \cellcolor{ResultRed!99!ResultYellow}0.7 & \cellcolor{ResultRed!94!ResultYellow}2.9 & \cellcolor{ResultRed!99!ResultYellow}0.5 & \cellcolor{ResultRed!99!ResultYellow}0.5 & \cellcolor{ResultRed!98!ResultYellow}1.2 & \cellcolor{ResultRed!99!ResultYellow}0.6 & \cellcolor{ResultRed!95!ResultYellow}2.4 & \cellcolor{ResultRed!100!ResultYellow}0.0 & \cellcolor{ResultRed!100!ResultYellow}0.0 & \cellcolor{ResultRed!99!ResultYellow}0.4 & \cellcolor{ResultRed!100!ResultYellow}0.0 & \cellcolor{ResultRed!86!ResultYellow}7.1 & \cellcolor{ResultRed!99!ResultYellow}0.3 & \cellcolor{ResultRed!97!ResultYellow}1.5 & \cellcolor{ResultRed!96!ResultYellow}2.2 & \cellcolor{ResultRed!99!ResultYellow}0.4 & \cellcolor{ResultRed!47!ResultYellow}878.0 \\
G3Reg & \cellcolor{ResultYellow!0!ResultGreen}\textbf{100.0} & \cellcolor{ResultYellow!16!ResultGreen}91.9 & \cellcolor{ResultYellow!0!ResultGreen}\textbf{100.0} & \cellcolor{ResultYellow!5!ResultGreen}97.7 & \cellcolor{ResultYellow!0!ResultGreen}\textbf{100.0} & \cellcolor{ResultYellow!6!ResultGreen}\textbf{96.8} & \cellcolor{ResultYellow!66!ResultGreen}67.2 & \cellcolor{ResultYellow!5!ResultGreen}\textbf{97.5} & \cellcolor{ResultYellow!0!ResultGreen}\textbf{100.0} & \cellcolor{ResultYellow!2!ResultGreen}\textbf{98.8} & \cellcolor{ResultYellow!83!ResultGreen}\textbf{58.3} & \cellcolor{ResultYellow!96!ResultGreen}\underline{52.0} & \cellcolor{ResultYellow!53!ResultGreen}\textbf{73.5} & \cellcolor{ResultYellow!2!ResultGreen}\textbf{99.2} & \cellcolor{ResultYellow!46!ResultGreen}\textbf{76.8} & \cellcolor{ResultYellow!33!ResultGreen}\textbf{83.4} & \cellcolor{ResultYellow!63!ResultGreen}68.6 & \cellcolor{ResultYellow!21!ResultGreen}\textbf{89.4} & \cellcolor{ResultYellow!2!ResultGreen}\textbf{99.2} & \cellcolor{ResultYellow!19!ResultGreen}\textbf{90.6} & \cellcolor{ResultYellow!0!ResultGreen}\textbf{96.5} \\
\addlinespace[2pt]
\multicolumn{22}{l}{\textit{Ours}} \\[-1pt]
\textbf{PARTE} & \cellcolor{ResultYellow!0!ResultGreen}\textbf{100.0} & \cellcolor{ResultYellow!0!ResultGreen}\textbf{100.0} & \cellcolor{ResultYellow!0!ResultGreen}\textbf{100.0} & \cellcolor{ResultYellow!0!ResultGreen}\textbf{100.0} & \cellcolor{ResultYellow!0!ResultGreen}\textbf{100.0} & \cellcolor{ResultYellow!14!ResultGreen}93.0 & \cellcolor{ResultYellow!28!ResultGreen}\textbf{85.9} & \cellcolor{ResultYellow!6!ResultGreen}\underline{97.0} & \cellcolor{ResultYellow!0!ResultGreen}\textbf{100.0} & \cellcolor{ResultYellow!4!ResultGreen}\underline{98.1} & \cellcolor{ResultYellow!88!ResultGreen}\underline{56.2} & \cellcolor{ResultYellow!90!ResultGreen}\textbf{55.2} & \cellcolor{ResultYellow!85!ResultGreen}\underline{57.5} & \cellcolor{ResultYellow!6!ResultGreen}\underline{96.8} & \cellcolor{ResultYellow!64!ResultGreen}\underline{68.1} & \cellcolor{ResultYellow!37!ResultGreen}\underline{81.4} & \cellcolor{ResultYellow!43!ResultGreen}\textbf{78.4} & \cellcolor{ResultYellow!33!ResultGreen}\underline{83.3} & \cellcolor{ResultYellow!3!ResultGreen}\underline{98.7} & \cellcolor{ResultYellow!26!ResultGreen}\underline{87.1} & \cellcolor{ResultYellow!6!ResultGreen}127.1 \\
\bottomrule
\end{tabular}%

}
\end{table*}

KITTI-LC provides a substantially more difficult setting by evaluating temporally separated loop closures over a wider range of relative motion.
Table~\ref{tab:kittilc-registration} therefore separates the results into $0$--$10$ m, $10$--$20$ m, and $20$--$30$ m translation ranges in addition to the overall result.

While several methods perform nearly perfectly at short range, their success rates decrease substantially as the displacement increases. 
PARTE remains among the strongest methods across the sequences and distance ranges, with particularly strong performance in the short- and medium-range cases.
At $20$--$30$ m, G3Reg achieves the highest success rate on four of the five sequences.
This is a favorable setting for G3Reg, whose frontend is specifically designed for outdoor LiDAR registration and exploits planes, lines, and object clusters that are common in structured urban environments~\cite{qiao2024g3reg}.

\textbf{Partial-to-full registration.}
We additionally evaluate registration on the RESSO partial-to-full benchmark, where a local RGB-D scan is aligned to a larger static reference scan.
This setting is challenging because the two point clouds can have substantially lower overlap and different scene context.

Table~\ref{tab:resso} reports the results across the seven evaluated scenes. 
PARTE achieves the highest overall success rate of $59.6\%$, followed by CLIPPER+ at $58.6\%$, MAC and VBReg at $57.6\%$, and TEASER++ at $56.6\%$.
The smaller margin compared with the main indoor and outdoor benchmarks is consistent with the plane-matching results in Table~\ref{tab:plane-matching-results}.
In partial-to-full registration, the context surrounding a corresponding plane can change substantially between scans, reducing the discriminative advantage of PCH as is evident in Table~\ref{tab:plane-matching-results}.
PARTE fails on 40 of the 99 registrations, of which 28 are also missed by every other method in Table~\ref{tab:resso}.
Among these common failures, 25 contain fewer than five geometrically consistent FPFH correspondences and 13 contain none.
Of the remaining PARTE failures, several result from an incorrect plane match that biases the selected clique toward outliers.

These cases expose a limitation of the joint point--plane formulation: confidence-weighted plane correspondences improve robustness when reliable, but incorrect associations can reinforce an erroneous consensus.
Nevertheless, PARTE achieves the highest overall success rate and succeeds on six pairs where both VBReg and TEASER++ fail.

\begin{table}[th]
\centering
\caption{Partial-to-full registration performance on the seven evaluated
RESSO scenes. We report success rate for each scene and overall, together with mean end-to-end runtime in milliseconds.}
\label{tab:resso}
\scriptsize
\setlength{\tabcolsep}{2pt}
\resizebox{\columnwidth}{!}{%
\begin{tabular}{lccccccccc}
\toprule
Method & \multicolumn{8}{c}{Success Rate (\%)} & Time \\
\cmidrule(lr){2-9}
 & 6a & 6b & 6c & 6d & 6f & 6g & 6h & All &  \\
\midrule
\multicolumn{10}{l}{\textit{Baseline}} \\[-1pt]
RANSAC-$10^{3}$ & \cellcolor{ResultRed!100!ResultYellow}0.0 & \cellcolor{ResultRed!68!ResultYellow}15.8 & \cellcolor{ResultRed!100!ResultYellow}0.0 & \cellcolor{ResultRed!73!ResultYellow}13.3 & \cellcolor{ResultRed!80!ResultYellow}10.0 & \cellcolor{ResultRed!100!ResultYellow}0.0 & \cellcolor{ResultRed!100!ResultYellow}0.0 & \cellcolor{ResultRed!88!ResultYellow}6.1 & \cellcolor{ResultYellow!63!ResultGreen}348.8 \\
RANSAC-$10^{5}$ & \cellcolor{ResultRed!54!ResultYellow}23.1 & \cellcolor{ResultRed!5!ResultYellow}47.4 & \cellcolor{ResultRed!26!ResultYellow}36.8 & \cellcolor{ResultYellow!93!ResultGreen}53.3 & \cellcolor{ResultRed!40!ResultYellow}30.0 & \cellcolor{ResultRed!100!ResultYellow}0.0 & \cellcolor{ResultRed!43!ResultYellow}28.6 & \cellcolor{ResultRed!31!ResultYellow}34.3 & \cellcolor{ResultYellow!66!ResultGreen}358.1 \\
RANSAC-$10^{7}$ & \cellcolor{ResultYellow!46!ResultGreen}\textbf{76.9} & \cellcolor{ResultRed!5!ResultYellow}47.4 & \cellcolor{ResultRed!16!ResultYellow}\underline{42.1} & \cellcolor{ResultYellow!53!ResultGreen}73.3 & \cellcolor{ResultYellow!40!ResultGreen}\textbf{80.0} & \cellcolor{ResultRed!100!ResultYellow}0.0 & \cellcolor{ResultYellow!86!ResultGreen}\textbf{57.1} & \cellcolor{ResultYellow!91!ResultGreen}54.5 & \cellcolor{ResultRed!67!ResultYellow}727.1 \\
FGR & \cellcolor{ResultRed!54!ResultYellow}23.1 & \cellcolor{ResultRed!58!ResultYellow}21.1 & \cellcolor{ResultRed!37!ResultYellow}31.6 & \cellcolor{ResultYellow!93!ResultGreen}53.3 & \cellcolor{ResultRed!60!ResultYellow}20.0 & \cellcolor{ResultRed!100!ResultYellow}0.0 & \cellcolor{ResultRed!43!ResultYellow}28.6 & \cellcolor{ResultRed!45!ResultYellow}27.3 & \cellcolor{ResultYellow!67!ResultGreen}362.7 \\
\addlinespace[1.5pt]
\multicolumn{10}{l}{\textit{Robust point-based methods}} \\[-1pt]
KISS-Matcher & \cellcolor{ResultYellow!77!ResultGreen}61.5 & \cellcolor{ResultRed!37!ResultYellow}31.6 & \cellcolor{ResultRed!37!ResultYellow}31.6 & \cellcolor{ResultYellow!53!ResultGreen}73.3 & \cellcolor{ResultYellow!80!ResultGreen}60.0 & \cellcolor{ResultRed!78!ResultYellow}11.1 & \cellcolor{ResultRed!43!ResultYellow}28.6 & \cellcolor{ResultRed!15!ResultYellow}42.4 & \cellcolor{ResultYellow!26!ResultGreen}\underline{213.5} \\
CLIPPER+ & \cellcolor{ResultYellow!46!ResultGreen}\textbf{76.9} & \cellcolor{ResultYellow!95!ResultGreen}\underline{52.6} & \cellcolor{ResultRed!16!ResultYellow}\underline{42.1} & \cellcolor{ResultYellow!40!ResultGreen}\underline{80.0} & \cellcolor{ResultYellow!60!ResultGreen}\underline{70.0} & \cellcolor{ResultRed!33!ResultYellow}\underline{33.3} & \cellcolor{ResultYellow!86!ResultGreen}\textbf{57.1} & \cellcolor{ResultYellow!83!ResultGreen}\underline{58.6} & \cellcolor{ResultRed!62!ResultYellow}707.9 \\
TEASER++ & \cellcolor{ResultYellow!62!ResultGreen}\underline{69.2} & \cellcolor{ResultRed!5!ResultYellow}47.4 & \cellcolor{ResultRed!5!ResultYellow}\textbf{47.4} & \cellcolor{ResultYellow!27!ResultGreen}\textbf{86.7} & \cellcolor{ResultYellow!60!ResultGreen}\underline{70.0} & \cellcolor{ResultRed!56!ResultYellow}22.2 & \cellcolor{ResultRed!0!ResultYellow}\underline{50.0} & \cellcolor{ResultYellow!87!ResultGreen}56.6 & \cellcolor{ResultYellow!73!ResultGreen}385.2 \\
QUATRO & \cellcolor{ResultRed!100!ResultYellow}0.0 & \cellcolor{ResultRed!100!ResultYellow}0.0 & \cellcolor{ResultRed!5!ResultYellow}\textbf{47.4} & \cellcolor{ResultRed!87!ResultYellow}6.7 & \cellcolor{ResultRed!100!ResultYellow}0.0 & \cellcolor{ResultRed!100!ResultYellow}0.0 & \cellcolor{ResultRed!100!ResultYellow}0.0 & \cellcolor{ResultRed!80!ResultYellow}10.1 & \cellcolor{ResultYellow!73!ResultGreen}385.1 \\
MAC & \cellcolor{ResultYellow!46!ResultGreen}\textbf{76.9} & \cellcolor{ResultYellow!95!ResultGreen}\underline{52.6} & \cellcolor{ResultRed!16!ResultYellow}\underline{42.1} & \cellcolor{ResultYellow!40!ResultGreen}\underline{80.0} & \cellcolor{ResultYellow!40!ResultGreen}\textbf{80.0} & \cellcolor{ResultRed!78!ResultYellow}11.1 & \cellcolor{ResultYellow!86!ResultGreen}\textbf{57.1} & \cellcolor{ResultYellow!85!ResultGreen}57.6 & \cellcolor{ResultRed!19!ResultYellow}549.9 \\
\addlinespace[1.5pt]
\multicolumn{10}{l}{\textit{Learning-based methods}} \\[-1pt]
PointDSC & \cellcolor{ResultYellow!92!ResultGreen}53.8 & \cellcolor{ResultRed!5!ResultYellow}47.4 & \cellcolor{ResultRed!26!ResultYellow}36.8 & \cellcolor{ResultYellow!53!ResultGreen}73.3 & \cellcolor{ResultYellow!80!ResultGreen}60.0 & \cellcolor{ResultRed!78!ResultYellow}11.1 & \cellcolor{ResultRed!14!ResultYellow}42.9 & \cellcolor{ResultRed!5!ResultYellow}47.5 & \cellcolor{ResultYellow!72!ResultGreen}380.3 \\
VBReg & \cellcolor{ResultYellow!46!ResultGreen}\textbf{76.9} & \cellcolor{ResultYellow!95!ResultGreen}\underline{52.6} & \cellcolor{ResultRed!16!ResultYellow}\underline{42.1} & \cellcolor{ResultYellow!40!ResultGreen}\underline{80.0} & \cellcolor{ResultYellow!40!ResultGreen}\textbf{80.0} & \cellcolor{ResultRed!78!ResultYellow}11.1 & \cellcolor{ResultYellow!86!ResultGreen}\textbf{57.1} & \cellcolor{ResultYellow!85!ResultGreen}57.6 & \cellcolor{ResultYellow!89!ResultGreen}443.7 \\
PREDATOR & \cellcolor{ResultRed!100!ResultYellow}0.0 & \cellcolor{ResultRed!58!ResultYellow}21.1 & \cellcolor{ResultRed!47!ResultYellow}26.3 & \cellcolor{ResultRed!87!ResultYellow}6.7 & \cellcolor{ResultRed!100!ResultYellow}0.0 & \cellcolor{ResultRed!100!ResultYellow}0.0 & \cellcolor{ResultRed!100!ResultYellow}0.0 & \cellcolor{ResultRed!80!ResultYellow}10.1 & \cellcolor{ResultYellow!75!ResultGreen}392.3 \\
\addlinespace[1.5pt]
\multicolumn{10}{l}{\textit{Segment- and plane-based methods}} \\[-1pt]
VPFBR & \cellcolor{ResultRed!38!ResultYellow}30.8 & \cellcolor{ResultRed!100!ResultYellow}0.0 & \cellcolor{ResultRed!89!ResultYellow}5.3 & \cellcolor{ResultRed!60!ResultYellow}20.0 & \cellcolor{ResultRed!100!ResultYellow}0.0 & \cellcolor{ResultRed!78!ResultYellow}11.1 & \cellcolor{ResultRed!86!ResultYellow}7.1 & \cellcolor{ResultRed!80!ResultYellow}10.1 & \cellcolor{ResultYellow!13!ResultGreen}\textbf{166.3} \\
PLADE & \cellcolor{ResultRed!100!ResultYellow}0.0 & \cellcolor{ResultRed!100!ResultYellow}0.0 & \cellcolor{ResultRed!79!ResultYellow}10.5 & \cellcolor{ResultRed!73!ResultYellow}13.3 & \cellcolor{ResultRed!100!ResultYellow}0.0 & \cellcolor{ResultRed!78!ResultYellow}11.1 & \cellcolor{ResultRed!100!ResultYellow}0.0 & \cellcolor{ResultRed!90!ResultYellow}5.1 & \cellcolor{ResultRed!100!ResultYellow}6617.7 \\
\addlinespace[1.5pt]
\multicolumn{10}{l}{\textit{Ours}} \\[-1pt]
\textbf{PARTE} & \cellcolor{ResultYellow!46!ResultGreen}\textbf{76.9} & \cellcolor{ResultYellow!84!ResultGreen}\textbf{57.9} & \cellcolor{ResultRed!16!ResultYellow}\underline{42.1} & \cellcolor{ResultYellow!67!ResultGreen}66.7 & \cellcolor{ResultYellow!60!ResultGreen}\underline{70.0} & \cellcolor{ResultYellow!89!ResultGreen}\textbf{55.6} & \cellcolor{ResultYellow!86!ResultGreen}\textbf{57.1} & \cellcolor{ResultYellow!81!ResultGreen}\textbf{59.6} & \cellcolor{ResultYellow!54!ResultGreen}315.8 \\
\bottomrule
\end{tabular}%

}
\end{table}

\textbf{Additional environments.}
We finally evaluate on ETH, which contains comparatively few planar patches and often only a dominant ground plane.
As in the indoor experiments, we find that the orientation of the normals used by the FPFH frontend has a substantial effect on the reported registration rate.
For example, TEASER++ improves from $72.7\%$ to $94.0\%$ when FPFH is computed using consistently oriented normals.
Similarly, MAC reports a registration recall of $36.1\%$ with FPFH on ETH~\cite{zhang20233d}; in our evaluation this increases to $57.7\%$ with MNN matching and further to $87.2\%$ when the FPFH normals are consistently oriented.
This makes ETH considerably less challenging once a stronger point frontend is used.

\begin{table}[th]
\centering
\caption{Registration performance on the ETH benchmark.
Success rate is reported for each sequence and overall, together with mean
end-to-end runtime. MAC$^\dagger$ encountered out-of-memory failures on
17 of the 713 registration pairs; these cases are counted as failures.}
\label{tab:outdoor_eth}
\scriptsize
\setlength{\tabcolsep}{2pt}
\resizebox{\columnwidth}{!}{%
\begin{tabular}{lcccccc}
\toprule
Method & \multicolumn{5}{c}{Success Rate (\%)} & Time \\
\cmidrule(lr){2-6}
 & \shortstack{Gazebo\\Summer} & \shortstack{Gazebo\\Winter} & \shortstack{Wood\\Autumn} & \shortstack{Wood\\Summer} & All &  \\
\midrule
\multicolumn{7}{l}{\textit{Baseline}} \\[-1pt]
RANSAC-$10^{3}$ & \cellcolor{ResultRed!47!ResultYellow}26.6 & \cellcolor{ResultRed!82!ResultYellow}9.0 & \cellcolor{ResultRed!93!ResultYellow}3.5 & \cellcolor{ResultRed!86!ResultYellow}7.2 & \cellcolor{ResultRed!75!ResultYellow}12.3 & \cellcolor{ResultYellow!61!ResultGreen}872.4 \\
RANSAC-$10^{5}$ & \cellcolor{ResultYellow!43!ResultGreen}78.3 & \cellcolor{ResultYellow!91!ResultGreen}54.3 & \cellcolor{ResultYellow!70!ResultGreen}65.2 & \cellcolor{ResultYellow!80!ResultGreen}60.0 & \cellcolor{ResultYellow!73!ResultGreen}63.3 & \cellcolor{ResultYellow!63!ResultGreen}892.9 \\
RANSAC-$10^{7}$ & \cellcolor{ResultYellow!3!ResultGreen}98.4 & \cellcolor{ResultYellow!45!ResultGreen}77.5 & \cellcolor{ResultYellow!9!ResultGreen}95.7 & \cellcolor{ResultYellow!8!ResultGreen}96.0 & \cellcolor{ResultYellow!22!ResultGreen}89.1 & \cellcolor{ResultRed!4!ResultYellow}1373.7 \\
FGR & \cellcolor{ResultYellow!78!ResultGreen}60.9 & \cellcolor{ResultRed!27!ResultYellow}36.7 & \cellcolor{ResultYellow!57!ResultGreen}71.3 & \cellcolor{ResultYellow!51!ResultGreen}74.4 & \cellcolor{ResultYellow!90!ResultGreen}55.1 & \cellcolor{ResultYellow!67!ResultGreen}935.1 \\
\addlinespace[1.5pt]
\multicolumn{7}{l}{\textit{Robust point-based methods}} \\[-1pt]
KISS-Matcher & \cellcolor{ResultYellow!10!ResultGreen}95.1 & \cellcolor{ResultYellow!35!ResultGreen}82.4 & \cellcolor{ResultYellow!9!ResultGreen}95.7 & \cellcolor{ResultYellow!13!ResultGreen}93.6 & \cellcolor{ResultYellow!20!ResultGreen}89.8 & \cellcolor{ResultYellow!6!ResultGreen}226.7 \\
CLIPPER+ & \cellcolor{ResultYellow!1!ResultGreen}\underline{99.5} & \cellcolor{ResultYellow!21!ResultGreen}89.3 & \cellcolor{ResultYellow!2!ResultGreen}\underline{99.1} & \cellcolor{ResultYellow!10!ResultGreen}95.2 & \cellcolor{ResultYellow!11!ResultGreen}\underline{94.5} & \cellcolor{ResultRed!100!ResultYellow}4266.4 \\
TEASER++ & \cellcolor{ResultYellow!0!ResultGreen}\textbf{100.0} & \cellcolor{ResultYellow!27!ResultGreen}86.5 & \cellcolor{ResultYellow!3!ResultGreen}98.3 & \cellcolor{ResultYellow!3!ResultGreen}\underline{98.4} & \cellcolor{ResultYellow!12!ResultGreen}94.0 & \cellcolor{ResultYellow!79!ResultGreen}1078.3 \\
QUATRO & \cellcolor{ResultYellow!25!ResultGreen}87.5 & \cellcolor{ResultYellow!10!ResultGreen}\underline{95.2} & \cellcolor{ResultYellow!63!ResultGreen}68.7 & \cellcolor{ResultYellow!37!ResultGreen}81.6 & \cellcolor{ResultYellow!27!ResultGreen}86.5 & \cellcolor{ResultRed!14!ResultYellow}1479.3 \\
MAC$^\dagger$ & \cellcolor{ResultYellow!3!ResultGreen}98.4 & \cellcolor{ResultYellow!47!ResultGreen}76.5 & \cellcolor{ResultYellow!12!ResultGreen}93.9 & \cellcolor{ResultYellow!21!ResultGreen}89.6 & \cellcolor{ResultYellow!26!ResultGreen}87.2 & \cellcolor{ResultRed!100!ResultYellow}7521.2 \\
\addlinespace[1.5pt]
\multicolumn{7}{l}{\textit{Learning-based methods}} \\[-1pt]
PointDSC & \cellcolor{ResultYellow!16!ResultGreen}91.8 & \cellcolor{ResultYellow!42!ResultGreen}79.2 & \cellcolor{ResultYellow!37!ResultGreen}81.7 & \cellcolor{ResultYellow!46!ResultGreen}76.8 & \cellcolor{ResultYellow!35!ResultGreen}82.5 & \cellcolor{ResultYellow!64!ResultGreen}905.0 \\
VBReg & \cellcolor{ResultYellow!7!ResultGreen}96.7 & \cellcolor{ResultYellow!21!ResultGreen}89.3 & \cellcolor{ResultYellow!21!ResultGreen}89.6 & \cellcolor{ResultYellow!30!ResultGreen}84.8 & \cellcolor{ResultYellow!19!ResultGreen}90.5 & \cellcolor{ResultYellow!74!ResultGreen}1015.4 \\
PREDATOR & \cellcolor{ResultYellow!93!ResultGreen}53.3 & \cellcolor{ResultYellow!72!ResultGreen}64.0 & \cellcolor{ResultYellow!57!ResultGreen}71.3 & \cellcolor{ResultYellow!54!ResultGreen}72.8 & \cellcolor{ResultYellow!72!ResultGreen}64.0 & \cellcolor{ResultRed!29!ResultYellow}1657.8 \\
\addlinespace[1.5pt]
\multicolumn{7}{l}{\textit{Segment- and plane-based methods}} \\[-1pt]
VPFBR & \cellcolor{ResultRed!33!ResultYellow}33.7 & \cellcolor{ResultRed!70!ResultYellow}14.9 & \cellcolor{ResultRed!91!ResultYellow}4.3 & \cellcolor{ResultRed!89!ResultYellow}5.6 & \cellcolor{ResultRed!67!ResultYellow}16.4 & \cellcolor{ResultYellow!0!ResultGreen}\textbf{125.9} \\
PLADE & \cellcolor{ResultRed!90!ResultYellow}4.9 & \cellcolor{ResultRed!88!ResultYellow}5.9 & \cellcolor{ResultRed!95!ResultYellow}2.6 & \cellcolor{ResultRed!94!ResultYellow}3.2 & \cellcolor{ResultRed!91!ResultYellow}4.6 & \cellcolor{ResultYellow!47!ResultGreen}708.9 \\
G3Reg & \cellcolor{ResultRed!40!ResultYellow}29.9 & \cellcolor{ResultRed!60!ResultYellow}20.1 & \cellcolor{ResultRed!4!ResultYellow}47.8 & \cellcolor{ResultYellow!96!ResultGreen}52.0 & \cellcolor{ResultRed!35!ResultYellow}32.7 & \cellcolor{ResultYellow!0!ResultGreen}\underline{161.9} \\
\addlinespace[1.5pt]
\multicolumn{7}{l}{\textit{Ours}} \\[-1pt]
\textbf{PARTE} & \cellcolor{ResultYellow!0!ResultGreen}\textbf{100.0} & \cellcolor{ResultYellow!2!ResultGreen}\textbf{99.0} & \cellcolor{ResultYellow!0!ResultGreen}\textbf{100.0} & \cellcolor{ResultYellow!0!ResultGreen}\textbf{100.0} & \cellcolor{ResultYellow!1!ResultGreen}\textbf{99.6} & \cellcolor{ResultYellow!24!ResultGreen}437.6 \\
\bottomrule
\end{tabular}%

}
\end{table}

Although ETH is largely unstructured and often contains only the ground plane as a reliable planar feature, the available plane correspondences still provide useful complementary constraints.
PARTE recovers 40 of the 43 registrations missed by TEASER++, with plane correspondences participating in 39 of these recoveries.
The ground plane is selected in all 39 cases, while only 12 contain an additional plane correspondence.
This shows that even a single reliable plane can provide useful complementary evidence when the point correspondences alone are insufficient.

Gazebo Winter provides an especially interesting example.
The relative motion is close to gravity-aligned, with a maximum tilt of $5.48^\circ$, making it well suited to the reduced rotational model exploited by QUATRO~\cite{lim2024quatropp}.
PARTE recovers 95 of the 96 registrations missed by QUATRO.
In 92 of these cases, QUATRO's point-only maximum clique contains incorrect correspondences, with outliers forming the majority in 36 cases.
PARTE uses the matched ground plane directly in the compatibility graph, where plane--point consistency removes point correspondences that disagree with the shared plane before transformation estimation.
Therefore, the ground plane provides information during correspondence selection, rather than only constraining the final pose.

Ablation studies of our algorithm for planar-patch extraction, plane weight selection, and runtime are presented in Appendices \ref{app:a} and \ref{app:b}.
Fig.~\ref{fig:overall-registration} summarizes performance across the four main benchmarks.
\textbf{PARTE achieves the highest mean success rate while maintaining low runtime, showing consistently strong performance across both indoor and outdoor} registration problems.

\subsection{Limitations}
The experiments show that reliable plane correspondences can substantially improve global registration, particularly when the point correspondence set contains few inliers.
At the same time, PARTE remains fundamentally a mixed point--plane method, and its behavior approaches that of the underlying point pipeline when few useful planes are available.
As discussed in Appendix~\ref{sec:runtime}, point descriptor computation and matching also remain the main computational bottleneck.

A second limitation is sensitivity to changes in the geometry and density surrounding a plane.
This is most evident on RESSO in Table~\ref{tab:plane-matching-results}, where partial-to-full registration can substantially change the context observed around corresponding planes and reduce the effectiveness of PCH.
Large density variations, particularly in LiDAR, can produce a similar effect.

\begin{figure}[t]
        \centering
        \includegraphics[width=\columnwidth]{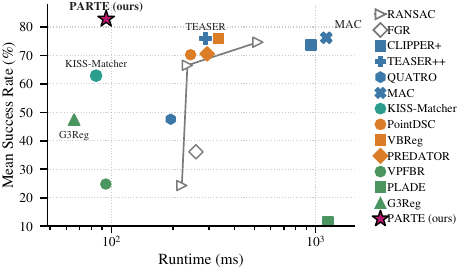}
\caption{Overall registration performance across 3DMatch, 3DLoMatch,
KITTI-10m, and KITTI-LC. Each point reports the mean success rate and runtime
across the four benchmarks, with each benchmark weighted equally.}
        \label{fig:overall-registration}
\end{figure}

\section{Conclusion}
\label{sec:conclusion}
We presented PARTE, a global point-cloud registration method that complements conventional point features with explicitly extracted and matched planar patches.
Our Plane Context Histogram (PCH) algorithm describes each plane using the surrounding geometry, enabling reliable plane correspondences when the patch itself is ambiguous.
These correspondences are combined with point matches during outlier rejection and transformation estimation, while the method naturally reduces to point-only registration when useful planes are unavailable.
Across diverse indoor RGB-D and outdoor LiDAR benchmarks, PARTE achieves consistently strong registration performance and the best overall success rate among the evaluated methods.

\noindent\textbf{Acknowledgment.} This research was supported by ARL SARA under Grant W911NF2420029.
The authors used a Large Language Model (OpenAI, GPT-5.6) for minor language editing; all technical content and results were developed and verified by the authors, who take full responsibility.

\bibliographystyle{IEEEtran}
\bibliography{refs}

\begin{appendices}

\section{Ablation Studies}
\label{app:a}

\begin{figure}[t]
    \centering
    \includegraphics[width=\columnwidth]{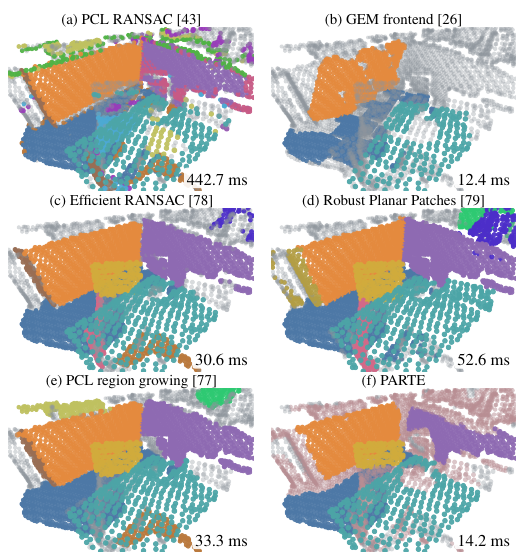}
\caption{Comparison of plane-extraction methods on a representative scene from the RedKitchen dataset.
Colored regions denote extracted planar patches. 
In PARTE, red points fail the initial planarity test and are excluded from region growing, leaving them available for point-feature matching. Reported times show plane-extraction runtime; the PARTE timing also includes KNN neighborhood search.}
\label{fig:plane-seg-comp}
\end{figure}

\textbf{Plane extraction.}
The plane-extraction method introduced in Sec.~\ref{sec:preprocessing} is designed specifically for the mixed point--plane registration pipeline rather than for maximizing segmentation quality.
During initialization, points whose neighborhoods do not satisfy the planarity conditions are excluded from region growing and remain available as point features.
As illustrated in Fig.~\ref{fig:plane-seg-comp}, this produces conservative planar patches while preserving edges, corners, and other non-planar regions.

\begin{figure}[t]
    \centering
    \includegraphics[width=\columnwidth]{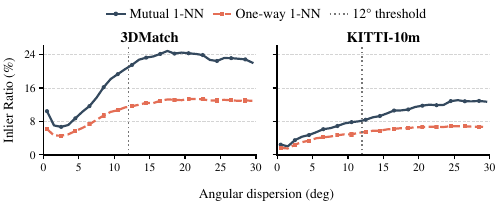}
\caption{FPFH inlier ratio as a function of local normal dispersion on 3DMatch and KITTI-10m. Points with larger normal dispersion are more likely to produce correct FPFH correspondences, supporting the $12^\circ$ threshold used to exclude such regions from plane growing.}
    \label{fig:plane-seg-dispersion}
\end{figure}

This early rejection also reduces the segmentation workload. On average, the initialization test excludes approximately $40\%$ of the points from region growing on 3DMatch, $50\%$ on KITTI-10m, and $65\%$ on ETH.
Fig.~\ref{fig:plane-seg-dispersion} shows that these rejected regions are also more useful for point matching: the FPFH inlier ratio generally increases with normal dispersion.
Therefore, the same test both avoids absorbing distinctive point features into planes and reduces unnecessary region-growing operations.
Table~\ref{tab:plane-seg-ablation} compares several alternative plane-extraction backends within the same registration pipeline.
\begin{table}[th]
\centering
\scriptsize
\caption{Effect of the plane-extraction backend on registration performance.
SR reports the success rate of the complete PARTE pipeline, while Time
reports the mean plane-extraction time per scan.}
\label{tab:plane-seg-ablation}
\setlength{\tabcolsep}{2pt}
\renewcommand{\arraystretch}{0.9}
\resizebox{\columnwidth}{!}{%
\begin{tabular}{lcc@{\hspace{10pt}}cc@{\hspace{10pt}}cc}
\toprule
\multirow{2}{*}{Plane extraction} & \multicolumn{2}{c}{3DMatch} & \multicolumn{2}{c}{3DLoMatch} & \multicolumn{2}{c}{KITTI-10m} \\
\cmidrule(lr){2-3}\cmidrule(lr){4-5}\cmidrule(lr){6-7}
 & SR (\%) & Time (ms) & SR (\%) & Time (ms) & SR (\%) & Time (ms) \\
\midrule
PCL RANSAC~\cite{fischler1981random} & \cellcolor{ResultYellow!43!ResultGreen}78.5 & \cellcolor{ResultRed!100!ResultYellow}254.4 & \cellcolor{ResultRed!45!ResultYellow}27.4 & \cellcolor{ResultRed!100!ResultYellow}251.3 & \cellcolor{ResultYellow!19!ResultGreen}90.3 & \cellcolor{ResultRed!100!ResultYellow}860.9 \\
PCL region growing~\cite{rabbani2006segmentation} & \cellcolor{ResultYellow!46!ResultGreen}76.8 & \cellcolor{ResultYellow!16!ResultGreen}18.9 & \cellcolor{ResultRed!46!ResultYellow}26.9 & \cellcolor{ResultYellow!16!ResultGreen}18.7 & \cellcolor{ResultYellow!5!ResultGreen}\underline{97.7} & \cellcolor{ResultYellow!10!ResultGreen}43.1 \\
Efficient RANSAC~\cite{schnabel2007efficient} & \cellcolor{ResultYellow!28!ResultGreen}\underline{85.8} & \cellcolor{ResultYellow!10!ResultGreen}13.3 & \cellcolor{ResultRed!14!ResultYellow}\underline{43.2} & \cellcolor{ResultYellow!10!ResultGreen}13.3 & \cellcolor{ResultYellow!19!ResultGreen}90.6 & \cellcolor{ResultYellow!13!ResultGreen}54.4 \\
Robust Planar Patches~\cite{araujo2020robust} & \cellcolor{ResultYellow!50!ResultGreen}75.2 & \cellcolor{ResultYellow!11!ResultGreen}14.2 & \cellcolor{ResultRed!48!ResultYellow}26.2 & \cellcolor{ResultYellow!11!ResultGreen}14.3 & \cellcolor{ResultYellow!20!ResultGreen}89.9 & \cellcolor{ResultYellow!5!ResultGreen}24.5 \\
GEM frontend~\cite{qiao2024g3reg} & \cellcolor{ResultYellow!33!ResultGreen}83.4 & \cellcolor{ResultYellow!0!ResultGreen}\textbf{2.9} & \cellcolor{ResultRed!22!ResultYellow}39.2 & \cellcolor{ResultYellow!0!ResultGreen}\textbf{2.8} & \cellcolor{ResultYellow!0!ResultGreen}\textbf{99.8} & \cellcolor{ResultYellow!0!ResultGreen}\textbf{6.8} \\
\textbf{PARTE} & \cellcolor{ResultYellow!21!ResultGreen}\textbf{89.4} & \cellcolor{ResultYellow!4!ResultGreen}\underline{7.5} & \cellcolor{ResultYellow!88!ResultGreen}\textbf{56.0} & \cellcolor{ResultYellow!4!ResultGreen}\underline{7.4} & \cellcolor{ResultYellow!0!ResultGreen}\textbf{99.8} & \cellcolor{ResultYellow!1!ResultGreen}\underline{12.0} \\
\bottomrule
\end{tabular}%

}
\end{table}

PARTE provides the strongest registration performance while remaining faster than most reliable segmentation baselines.
For completeness, the reported PARTE segmentation time includes KNN construction, although these neighborhoods are reused elsewhere in the pipeline.

\textbf{Weighted correspondence selection.}
We next evaluate how plane correspondences should contribute during clique selection.
We compare the standard unweighted maximum clique against two weighted variants.
The first assigns the same weight to every plane correspondence, while the second scales each plane weight using its PCH confidence $w_j$.
In both cases, a weight multiplier controls the overall influence of plane correspondences relative to unit-weight point matches.

\begin{figure}[t]
    \centering
    \includegraphics[width=\columnwidth]{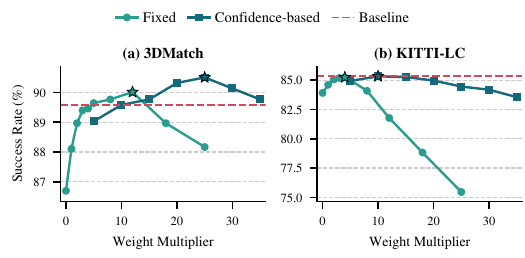}
    \caption{Effect of plane-correspondence weighting on registration success.
    Fixed weighting assigns the same weight to all plane matches, while
    confidence-based weighting scales their contribution using PCH confidence.
}
    \label{fig:plane-weight-ablation}
\end{figure}

Fig.~\ref{fig:plane-weight-ablation} shows the effect of varying the weight multiplier on 3DMatch and KITTI-LC.
Confidence-based weighting is considerably less sensitive to the multiplier, maintaining strong performance over a broader range.
This supports using the PCH confidence to distinguish reliable plane matches from ambiguous ones during clique selection.

\section{Runtime Analysis}
\label{app:b}

\label{sec:runtime}
\begin{figure}[t]
    \centering
    \includegraphics[width=\columnwidth]{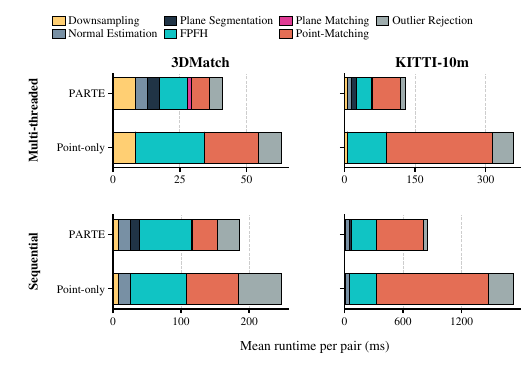}
    \caption{Runtime breakdown of PARTE and the point-only variant on 3DMatch
    and KITTI-10m under multithreaded and sequential execution. Plane extraction
    and matching add overhead, which is offset by lower point-feature,
    matching, and outlier-rejection costs.}
    \label{fig:runtime-breakdown}
\end{figure}

To understand where PARTE spends computation, we compare its runtime breakdown against a point-only variant.
The point-only variant uses the same downsampling, normal estimation, FPFH matching, and robust registration pipeline, but does not extract or match planes.
Instead, FPFH descriptors are computed over the full downsampled point cloud.
Fig.~\ref{fig:runtime-breakdown} shows the average runtime on 3DMatch and KITTI-10m under both multithreaded and sequential execution.
Overall, separating planar points and matching them as patches can substantially reduce the total runtime by lowering the cost of point descriptor computation, matching, and outlier rejection.
Nevertheless, PARTE's runtime remains dominated by the conventional point-matching stage.
\end{appendices}
\end{document}